\documentclass{mlrpreprint}

\usepackage{amsmath}
\usepackage{enumerate}
\usepackage{algorithm}
\usepackage{algpseudocode}
\usepackage{amsfonts}
\usepackage{amsthm}
\usepackage{cleveref}
\usepackage{diagbox}
\usepackage{colortbl}
\usepackage{amssymb}
\usepackage{xspace}
\usepackage{wrapfig}
\usepackage{adjustbox}
\usepackage{tabularx}
\usepackage{booktabs}
\usepackage{mathtools}
\usepackage{tikz}
\usepackage{enumitem}
\usepackage{silence}
\usepackage{dsfont}
\usepackage[table]{xcolor}
\usepackage[dvipsnames]{xcolor}
\usepackage{multirow}
\usepackage{makecell}
\usepackage{xfakebold}
\usepackage{listings}
\usepackage{nicefrac}
\usepackage{url}
\usepackage{fontawesome5}

\usepackage{amsmath,amsfonts,bm}

\def\eqref#1{equation~\ref{#1}}

\def\1{\bm{1}}

\DeclareMathAlphabet{\mathsfit}{\encodingdefault}{\sfdefault}{m}{sl}
\SetMathAlphabet{\mathsfit}{bold}{\encodingdefault}{\sfdefault}{bx}{n}

\definecolor{textgray}{HTML}{6E6E73}
\usetikzlibrary{positioning, calc}
\usetikzlibrary{decorations.pathmorphing}

\makeatletter
\patchcmd{\wrong@fontshape}{\@gobbletwo}{}{}{}
\makeatother
\numberwithin{equation}{section}
\makeatletter
\AtBeginDocument{
  \urlstyle{sf}
  
}
\makeatother

\definecolor{light}{RGB}{125, 125, 125}
\crefname{tcb@cnt@pbox}{code}{code}
\Crefname{tcb@cnt@pbox}{Code}{Code}
\crefname{assumption}{assumption}{assumption}
\Crefname{assumption}{Assumption}{Assumptions}

\newtcolorbox[auto counter]{pbox}[2][]{
  colback=white,
  title=Code~\thetcbcounter: #2,
  #1,fonttitle=\sffamily,
  fontupper=\sffamily,
  arc=2pt,
  colframe=bgcolor,
  coltitle=fgcolor,
  colbacktitle=bgcolor,
  toptitle=0.25cm,
  bottomtitle=0.125cm
}

\definecolor{cverbbg}{gray}{0.90}

\newcommand*\methodname{NF-CoT}

\newcommand{\pos}[1]{\textcolor{green!60!black}{\scriptsize +#1}}
\newcommand{\decr}[1]{\textcolor{red}{\scriptsize -#1}}

\definecolor{latentbg}{RGB}{255, 244, 214}
\definecolor{headerbg}{RGB}{245, 224, 179}
\definecolor{sectionbg}{RGB}{232,224,255}
\definecolor{subsectionbg}{RGB}{238,238,238}
\definecolor{cocoblue}{HTML}{BFE6F8}
\definecolor{cocoblueframe}{HTML}{2B8FBD}
\definecolor{cocogreen}{HTML}{DDF3D8}
\definecolor{cocogreenframe}{HTML}{3B8F49}
\definecolor{cocoyellow}{HTML}{FFE7B3}
\definecolor{cocoyellowframe}{HTML}{B9821A}
\definecolor{cocopurple}{HTML}{D8D6EE}
\definecolor{cocopurpleframe}{HTML}{67639B}
\definecolor{cocored}{HTML}{F8D4CE}
\definecolor{cocogray}{HTML}{F5F6F8}
\definecolor{codebg}{HTML}{FAFAFB}

\lstdefinestyle{cocopystyle}{
  basicstyle=\ttfamily\footnotesize,
  breaklines=true,
  columns=fullflexible,
  keepspaces=true,
  showstringspaces=false,
  frame=single,
  rulecolor=\color{black!12},
  backgroundcolor=\color{codebg},
  xleftmargin=0.5em,
  xrightmargin=0.5em,
  aboveskip=0.5em,
  belowskip=0.5em
}

\newtcolorbox{cocotaskbox}[1]{
  enhanced,
  breakable,
  colback=cocogray,
  colframe=cocopurpleframe,
  boxrule=0.7pt,
  arc=2mm,
  left=1mm,
  right=1mm,
  top=1mm,
  bottom=1mm,
  title=\textbf{#1},
  coltitle=black,
  colbacktitle=cocopurple,
  fonttitle=\normalsize,
}

\newtcolorbox{latentbox}[1]{
  enhanced,
  breakable,
  colback=cocoyellow!28,
  colframe=cocoyellowframe,
  boxrule=0.6pt,
  arc=2mm,
  left=1mm,
  right=1mm,
  top=1mm,
  bottom=1mm,
  title=\textbf{Decoded Latent CoT: #1},
  coltitle=black,
  colbacktitle=cocoyellow,
}

\newtcolorbox{answerbox}[1]{
  enhanced,
  breakable,
  colback=cocogreen!30,
  colframe=cocogreenframe,
  boxrule=0.6pt,
  arc=2mm,
  left=1mm,
  right=1mm,
  top=1mm,
  bottom=1mm,
  title=\textbf{Generated Answer: #1},
  coltitle=black,
  colbacktitle=cocogreen,
}

\usepackage{xcolor}

\crefname{section}{Sec.}{Secs.}
\Crefname{section}{Section}{Sections}
\crefname{subsection}{Sec.}{Secs.}
\Crefname{subsection}{Section}{Sections}
\crefname{figure}{Fig.}{Figs.}
\Crefname{figure}{Figure}{Figures}
\crefname{table}{Table}{Tables}
\Crefname{table}{Table}{Tables}
\crefname{equation}{Eq.}{Eqs.}
\Crefname{equation}{Equation}{Equations}
\crefname{algorithm}{Alg.}{Algs.}
\Crefname{algorithm}{Algorithm}{Algorithms}
\crefname{appendix}{Appendix}{Appendices}
\Crefname{appendix}{Appendix}{Appendices}

\title{Latent Reasoning with Normalizing Flows}

\author[*,\dagger]{Guancheng Tu}
\author[*,\dagger]{Xiangjun Fu}
\author[\dagger]{Suhao Yu}
\author[\dagger]{Yao Tang}
\author[\ddagger]{Haoqiang Kang}
\author[\ddagger]{Lianhui Qin}
\author[\S]{Yizhe Zhang}
\author[\dagger]{Jiatao Gu}

\affiliation[\dagger]{University of Pennsylvania}
\affiliation[\ddagger]{UC San Diego}
\affiliation[\S]{Meta}

\contribution[*]{Equal contribution}

\abstract{
Large language models often improve reasoning by generating explicit chain-of-thought (CoT), demonstrating the importance of intermediate computation. However, textual CoT forces this computation through a discrete, serial, and communication-oriented token stream: each reasoning step must be verbalized before the model can proceed, even when the underlying update is semantic, uncertain, or only partially formed. Latent reasoning offers a higher-bandwidth alternative by performing intermediate computation in compact continuous states before committing to text. Yet existing latent-reasoning methods often sacrifice key advantages that make CoT effective in autoregressive language models, including native left-to-right generation, probabilistic sampling, compatibility with KV-cache decoding, and tractable likelihood estimation.
We propose \methodname{}, a latent reasoning framework that preserves these advantages by modeling continuous thoughts with normalizing flows. \methodname{} instantiates a TARFlow-style normalizing flow inside the LLM backbone, defining a tractable probability model over compact continuous thoughts distilled from explicit CoT. Continuous-thought positions are generated by an NF head, while text positions are generated by the standard LM head within the same causal stream. This design provides exact likelihoods for latent thoughts, enables probabilistic left-to-right decoding with the original KV cache, and supports direct policy-gradient optimization in the latent reasoning space. On code-generation benchmarks, \methodname{} improves pass rates over explicit-CoT and prior latent-reasoning baselines while substantially reducing intermediate-reasoning cost.
}

\correspondence{\{tgc, xjf, jgu32\}@seas.upenn.edu}
\website{\url{https://nf-cot.vercel.app}}

\begin{document}

\maketitle

\section{Introduction}

\begin{wrapfigure}{r}{0.45\textwidth}
    \vspace{-1.5em}
    \centering
    \includegraphics[width=0.45\textwidth]{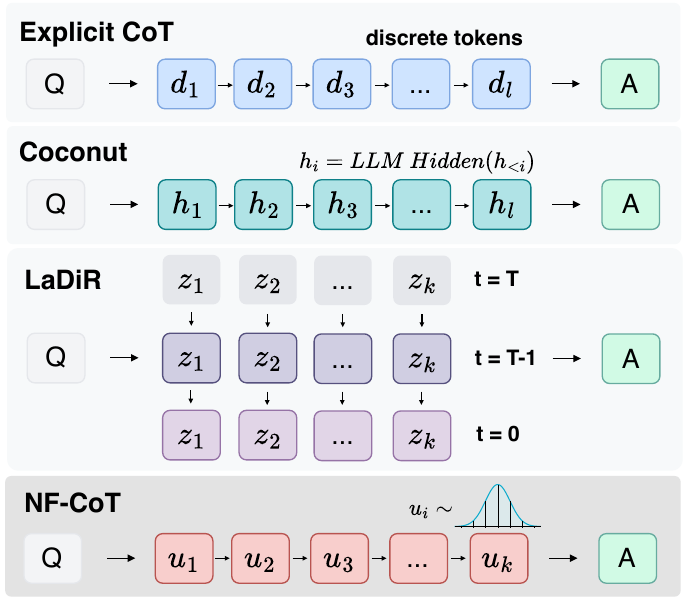}
    \caption{
        \textbf{Four paradigms for chain-of-thought reasoning.}
        \textbf{Explicit CoT}: \emph{discrete} text tokens.
        \textbf{Coconut}: \emph{deterministic} hidden states.
        \textbf{LaDiR}: \emph{iteratively} denoised latents using diffusion.
        \textbf{\methodname{} (ours)}: AR-sampled continuous thoughts.
    }
    \label{fig:teaser}
    \vspace{-1em}
\end{wrapfigure}
Chain-of-thought (CoT) prompting has become a standard chain-of-thoughtway to elicit reasoning in large language models (LLMs), improving performance by generating intermediate steps before the final answer~\citep{wei2022chain,kojima2022large,nye2021show}. One way to understand its effectiveness is that CoT introduces intermediate reasoning variables between the prompt and the answer, turning a direct input-output prediction into a conditioned prediction through a sampled reasoning path~\citep{zelikman2022star}. In explicit CoT, these variables are represented as text tokens, making them naturally autoregressive, probabilistic, and likelihood-scored under the LLM. Yet text is a verbose, low-information-density medium for thought, making long reasoning costly and tying intermediate computation to surface forms~\citep{hao2024coconut}.

This inefficiency has motivated latent CoT methods that replace textual reasoning traces with continuous or soft embedding states. Some methods keep reasoning close to the LLM by recycling hidden states or token-embedding mixtures as latent thoughts~\citep{hao2024coconut,zhang2026soft,tang2026multiplex}; they are efficient, but remain local and tied to the model's token or embedding space. Others model stochastic continuous latents with diffusion models~\citep{ladir2025,ladirl2026}, but require iterative denoising and lack native left-to-right likelihood modeling. The challenge is to learn latent CoT without giving up the native sampling, scoring, and decoding interface of language modeling in current LLMs.

In this work, we address this gap by proposing \textsc{\methodname{}}, a latent reasoning framework that gives continuous CoT the same modeling status as language tokens. \methodname{} places a scalable normalizing flow~\citep{zhai2024tarflow,gu2025starflow} inside the LLM's causal stream, using the LLM backbone as the shared autoregressive model for continuous thoughts and text answers. During training, \methodname{} turns explicit CoT supervision into continuous thoughts and optimizes an end-to-end likelihood objective over both latent reasoning and answers. The resulting tractable likelihood also enables policy-gradient optimization~\citep{deepseek2025r1} directly in the continuous reasoning space. At inference, \methodname{} samples continuous thoughts left-to-right and decodes answers in a single causal pass. Together, this preserves the sampling, decoding, and likelihood interface of explicit CoT in a compact continuous space, at lower token cost.

We evaluate \methodname{} on MBPP, MBPP+, HumanEval, HumanEval+, and LiveCodeBench~v6. \methodname{} improves pass rates over explicit-CoT and prior latent-reasoning baselines while substantially reducing intermediate-reasoning cost. We further study likelihood-based sampling and policy-gradient optimization in the continuous reasoning space.
To this end, our contributions are threefold:
\begin{itemize}
    \item A likelihood-based formulation of continuous CoT that preserves the sampling, scoring, and decoding interface of explicit CoT;
    \item A novel architecture that jointly models latent thoughts and text answers with NF and LM heads over a shared LLM backbone;
    \item A unified optimization interface for continuous thoughts, enabling supervised likelihood training and policy-gradient refinement.
\end{itemize}

\section{Preliminaries}
\label{sec:prelim}

\subsection{Chain-of-Thought Reasoning}
\label{sec:prelim-cot}
We consider supervised reasoning examples with a prompt $q$, a target answer sequence $x_{1:N}$, and, when available, an explicit chain-of-thought (CoT) trace $d_{1:L}$. A probabilistic view treats the trace as an intermediate variable:
\begin{equation}
p_\theta(x_{1:N}\mid q)
=
\sum_{d_{1:L}}
p_\theta(d_{1:L}\mid q)\,
p_\theta(x_{1:N}\mid q,d_{1:L}).
\label{eq.cot}
\end{equation}
Since $d_{1:L}$ is a text sequence, it can be generated, sampled, and scored with the same autoregressive language-modeling interface as the answer. This is convenient but forces reasoning into verbose natural-language tokens at high token cost.

\subsection{Continuous Chain-of-Thought}
\label{sec:prelim-continuous-cot}

Continuous CoT replaces text rationales with continuous codes $e_{1:K}\in\mathbb{R}^{K\times D}$ of length $K$ and dimension $D$, which play the role of the intermediate variable in \cref{eq.cot}.

Coconut~\citep{hao2024coconut} distills reasoning into hidden states fed back autoregressively, but the resulting thoughts are deterministic and do not define a distribution over reasoning paths. Another approach is to first learn a continuous code space with a VAE: an encoder maps $d_{1:L}$ together with $K$ learnable queries to $q_\phi(e_{1:K}\mid d_{1:L})$, trained with
\begin{equation}
\mathcal{L}_{\mathrm{VAE}}
= \mathbb{E}_{q_\phi(e\mid d)}\left[-\log p_\psi(d\mid e)\right]
+ \beta\,\mathrm{KL}\!\left(q_\phi(e\mid d)\,\|\,\mathcal{N}(0,I)\right).
\end{equation}
After pretraining, a separate generative model (e.g., diffusion~\citep{ladir2025}) is trained for $p_\theta(e_{1:K}\mid q)$. This gives a stochastic continuous representation but separates latent-thought generation from the native autoregressive likelihood interface used by explicit CoT.

\subsection{Scalable Normalizing Flows}
\label{sec:prelim-flows}

Normalizing flows~\citep{dinh2014nice,dinh2016density} map data $y$ to a simple base variable $z\sim\mathcal{N}(0,I)$ through an invertible network. For a conditional flow with context $c$,
\begin{equation}
z = f_\theta^{-1}(y;c), \qquad
p_\theta(y\mid c) = p(z)\left|\det J_{f_\theta^{-1}}(y;c)\right|,
\end{equation}
giving exact likelihood evaluation and direct sampling. Autoregressive flows~\citep{kingma2016improved,papamakarios2017maf} transform each position using only the context and previous positions, yielding a triangular Jacobian compatible with causal sequence modeling. Scalable variants such as TarFlow~\citep{zhai2024tarflow} and STARFlow~\citep{gu2025starflow} use Transformer blocks for high-dimensional structured latents. We build on this family to model continuous CoT with LLMs.

\begin{figure*}[t]
  \centering
  \includegraphics[width=\textwidth]{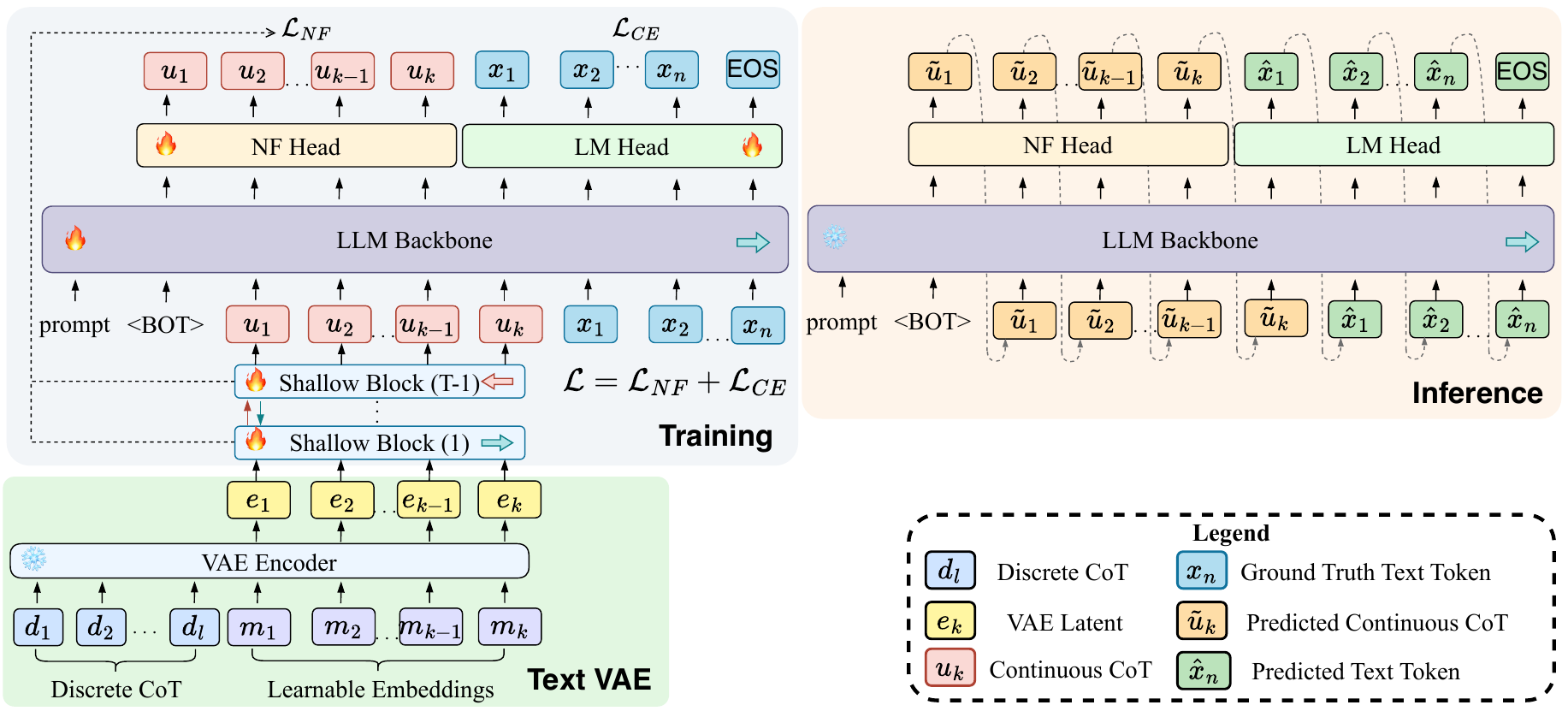}
  \caption{\textbf{Training and inference pipeline of \methodname{}.}
    Training uses a frozen VAE encoder and shallow flow blocks to convert
    discrete-CoT rationales into continuous-CoT states $u_{1:k}$. A shared LLM
    then consumes the prompt, \texttt{<BOT>}, continuous-CoT states, and
    ground-truth text tokens $x_{1:n}$ in one causal pass, jointly optimizing
    the flow NLL and answer cross-entropy. At inference, the VAE branch and
    shallow blocks are skipped: the LLM samples predicted continuous-CoT states
    $\tilde{u}_{1:k}$ left-to-right and continues decoding predicted text
    tokens $\hat{x}_{1:\hat n}$ with the same KV cache, without recomputing the prompt or CoT prefix.}
  \label{fig:training}
\end{figure*}

\section{Method}
\label{sec:method}

\subsection{Formulation}
\label{sec:method-formulation}

We start from the continuous CoT target $e_{1:K}$ defined in~\cref{sec:prelim-continuous-cot}. Since this encoder-derived space is optimized for text compression rather than autoregressive generation, its prompt-conditioned distribution can be difficult to model directly. Prior work thus turns to diffusion models~\citep{ladir2025}, trading expressive latent modeling for iterative sampling and a less direct likelihood interface.

\methodname{} takes a different route. We introduce an LLM-facing continuous thought space designed to make continuous thoughts easy to sample and score autoregressively. Following STARFlow~\citep{gu2025starflow}, we model continuous thought tokens $u_{1:K}$ with a causal Gaussian density:
\begin{equation}
\label{eq:u-ar-density}
p_\theta(u_{1:K} \mid q)
= \prod_{i=1}^K\mathcal{N}\!\left(
u_i;\,
\mu_\theta(q,u_{<i}),
\mathrm{diag}\!\left(\sigma_\theta^2(q,u_{<i})\right)
\right).
\end{equation}
where $\mu_\theta$ and $\sigma_\theta$ are causal functions of the prompt and previous thoughts. This space supports natural left-to-right sampling with tractable likelihood.

The two spaces are connected through an invertible transformation
$u_{1:K}=F_\theta(e_{1:K};q)$, implemented as a stack of shallow autoregressive
flow blocks following \citet{gu2025starflow}. Since $F_\theta$ is invertible,
$u_{1:K}$ is information-equivalent to $e_{1:K}$ rather than a lossy
compression. The full likelihood of the original continuous CoT target can
therefore be written through the reparameterized thought space:
\begin{equation}
\label{eq:e-likelihood-through-u}
\log p_\theta(e_{1:K}\mid q)
= \log p_\theta(u_{1:K}\mid q)
+ \log\left|\det J_{F_\theta}(e_{1:K};q)\right|.
\end{equation}
This allows the shallow flow blocks and the autoregressive density over
$u_{1:K}$ to be trained end-to-end under an exact likelihood. Thus,
\methodname{} learns from the same continuous CoT targets as prior latent
methods, but performs modeling and generation in an information-equivalent
thought space that is easier to sample and score autoregressively.

\subsection{Architecture}
\label{sec:method-architecture}

Architecturally, \methodname{} contains two flow components, as shown in \cref{fig:training}: shallow flow blocks for reparameterizing $e$ into $u$, followed by an LLM-parameterized deep autoregressive flow over $u$. The shallow blocks implement the invertible map $F_\theta:e_{1:K}\mapsto u_{1:K}$ and are trained from scratch with identity initialization ($F_\theta \approx I$). Because they are causal affine flows, their Jacobian is triangular and their log-determinant is tractable, contributing to the likelihood in \cref{eq:e-likelihood-through-u}.

The deep flow is implemented within the LLM backbone. Continuous thoughts $u_{1:K}$ are projected into the token embedding dimension and processed in the same causal stream as answer tokens, in the spirit of interleaved continuous-discrete generation~\citep{shen2026starflow2}. At continuous-thought positions, an NF head maps the hidden state to the parameters $\mu_\theta(q,u_{<i})$ and $\sigma_\theta(q,u_{<i})$ of the conditional density in \cref{eq:u-ar-density}. At answer positions, the standard LM head maps the hidden state to token logits. Thus, latent reasoning and answer generation share one causal backbone with separate heads for continuous and discrete variables. This unified design is the main architecture of \methodname{}; we report a dual-path variant, separating the latent-flow and answer-generation paths, as an ablation.

\subsection{Learning}
\label{sec:method-learning}

\paragraph{Supervision.}
We train \methodname{} from supervised reasoning examples with prompt $q$, explicit CoT trace $d_{1:L}$ when available, and answer $x_{1:N}$. As described in \cref{sec:prelim-continuous-cot}, we first pretrain a CoT encoder, freeze it, and use its posterior mean as the continuous CoT target $e_{1:K}$. The trainable model maps this target to the LLM-facing thought sequence $u_{1:K}=F_\theta(e_{1:K};q)$ and learns to model both the latent reasoning path and the final answer.

\paragraph{Unified likelihood objective.}
We train latent reasoning and answer generation with a single supervised objective. The flow term assigns exact likelihood to the original continuous CoT target, $\mathcal{L}_{\mathrm{flow}}=-\log p_\theta(e_{1:K}\mid q)$, computed through the reparameterized thought space using \cref{eq:e-likelihood-through-u}. The text term is the standard autoregressive likelihood conditioned on the continuous thoughts, $\mathcal{L}_{\mathrm{text}}=-\sum_{j=1}^{N}\log p_\theta(x_j\mid q,u_{1:K},x_{<j})$. The final objective is
\begin{equation}
\label{eq:sup-loss}
\mathcal{L}_{\mathrm{sup}}
=
\lambda_{\mathrm{flow}}\mathcal{L}_{\mathrm{flow}}
+
\lambda_{\mathrm{text}}\mathcal{L}_{\mathrm{text}}.
\end{equation}
Both terms are likelihoods under the same causal LLM: the flow head scores continuous thoughts, while the LM head scores answer tokens.

\paragraph{Training curriculum.}
We use a two-stage curriculum. First, we freeze the LLM backbone and train only the shallow flow blocks and continuous-thought projection layers with the same supervised objective in \cref{eq:sup-loss}, aligning the $e\!\to\!u$ reparameterization with the frozen LLM space. We then unfreeze all parameters and continue end-to-end training with the full objective. 

Following the practice in TarFlow~\citep{zhai2024tarflow}, we add small Gaussian noise to $e_{1:K}$ before computing the flow loss. We find that this improves robustness to deterministic target codes.

\subsection{Inference}
\label{sec:method-inference}

At inference time, \methodname{} does not run the shallow flow blocks $F_\theta$, which are only needed during training. Instead, inference proceeds directly in $u$-space: given a prompt $q$, the model samples continuous thoughts left-to-right from the autoregressive density in \cref{eq:u-ar-density}.
This sampling is stochastic, analogous to token sampling in a standard LLM, rather than a deterministic hidden-state update. After sampling $\tilde u_{1:K}$, the model switches to the LM head and continues answer generation in the same causal stream. Because continuous thoughts and answer tokens share the LLM backbone, the KV cache built during thought sampling is reused for answer generation at no extra cost.

This inference procedure differs from prior latent-CoT methods in two ways. First, \methodname{} samples from an explicit likelihood model over continuous thoughts rather than deterministically recycling hidden states. Second, it avoids diffusion-style iterative latent generation: although training uses the invertible map between $e$ and $u$, test-time reasoning happens directly in the information-equivalent $u$-space, which is shaped for autoregressive sampling and decoding.

\subsection{Reinforcement Learning}
\label{sec:method-rl}

After supervised training, we further refine \methodname{} with reinforcement learning. During this stage, we freeze the shallow flow blocks and optimize the policy directly in the LLM-facing thought space $u$. For a sampled thought-answer trajectory $(\tilde u_{1:K},\hat x_{1:\hat N})$, the policy likelihood decomposes as
\begin{equation}
\label{eq:joint-policy}
\log\pi_\theta(\tilde u,\hat x\mid q)
= \log p_\theta(\tilde u_{1:K}\mid q)
+ \log p_\theta(\hat x_{1:\hat N}\mid q,\tilde u_{1:K}).
\end{equation}
The first term is the Gaussian autoregressive likelihood from the NF head, and the second term is the token likelihood from the LM head. This makes policy-gradient optimization directly applicable to both continuous thoughts and text answers.

In our experiments, we use a GRPO-style objective over groups of sampled trajectories. Given a reward $R(q,\hat x)$, such as unit-test feedback for code generation, we compute a group-normalized advantage $A(q,\tilde u,\hat x)$ and update the policy via the policy gradient
\begin{equation}
\label{eq:rl-loss}
\nabla_\theta J_{\mathrm{RL}}
= \mathbb{E}\left[
A(q,\tilde u,\hat x)\,\nabla_\theta\log \pi_\theta(\tilde u,\hat x\mid q)
\right].
\end{equation}
Since the shallow blocks are frozen, this refinement is equivalent to applying policy gradients to the Gaussian likelihood over $u$ together with the token likelihood over answers in a single update.

\section{Experiments}
\label{sec:experiments}

We evaluate \methodname{} on code generation, a reasoning-intensive task with
standard correctness metrics. We compare with strong baselines, including
autoregressive CoT models, diffusion-based latent reasoners, and recent
continuous-reasoning methods. All methods use the same setup with
pass@1 and pass@k on multiple benchmarks.






\definecolor{sectionbg}{RGB}{229,222,255}      
\definecolor{subsectionbg}{RGB}{238,238,238}   
\definecolor{latentbg}{RGB}{255,245,218}       


\newcommand{\scorecell}[2]{\makebox[2.4em][r]{#1}\makebox[2.2em][l]{#2}}
\newcommand{\scoreonly}[1]{\makebox[2.4em][r]{#1}\makebox[2.2em][l]{}}

\begin{table*}[t]
\centering
\small
\caption{Pass@1 results on code generation benchmarks. Latent reasoning methods are highlighted in \colorbox{latentbg}{\strut light yellow}. Improvements over the base model are shown in \textcolor{green!60!black}{green}, and decreases in \textcolor{red}{red}. Best results across all reported methods are in \textbf{bold} and second-best results are \underline{underlined}.}
\resizebox{\textwidth}{!}{%
\begin{tabular}{l l c c c c c c}
\toprule
Model / Method & Base Model & MBPP & MBPP+ & HumanEval & HumanEval+ & LCB v6 & Avg. \\
\midrule

\rowcolor{sectionbg}
\multicolumn{8}{c}{\textbf{Open-Sourced Models}} \\
\rowcolor{subsectionbg}
\multicolumn{8}{c}{Autoregressive Coding Models} \\

Qwen 2.5 Coder* & Qwen2.5-7B 
& \scoreonly{75.9} & \scoreonly{62.9} & \scoreonly{66.5} & \scoreonly{60.4} & \scoreonly{26.8} & \scoreonly{59.9} \\

OpenCoder* & from scratch (8B) 
& \scoreonly{79.9} & \scoreonly{\underline{70.4}} & \scoreonly{66.5} & \scoreonly{63.4} & \scoreonly{29.6} & \scoreonly{62.0} \\

OlympicCoder* & Qwen2.5-7B 
& \scoreonly{80.0} & \scoreonly{66.4} & \scoreonly{82.1} & \scoreonly{76.9} & \scoreonly{\underline{37.3}} & \scoreonly{\underline{68.5}} \\

Seed-Coder* & from scratch 
& \scoreonly{\underline{82.0}} & \scoreonly{69.0} & \scoreonly{77.4} & \scoreonly{68.3} & \scoreonly{28.4} & \scoreonly{65.0} \\

\midrule
\rowcolor{subsectionbg}
\multicolumn{8}{c}{Diffusion Language Models} \\

Dream* & Qwen2.5-7B 
& \scoreonly{68.7} & \scoreonly{57.4} & \scoreonly{56.7} & \scoreonly{50.0} & \scoreonly{18.6} & \scoreonly{--} \\

LLaDA* & from scratch (8B) 
& \scoreonly{50.1} & \scoreonly{42.1} & \scoreonly{35.4} & \scoreonly{30.5} & \scoreonly{12.4} & \scoreonly{--} \\

Diffu-Coder* & Qwen2.5-Coder-7B 
& \scoreonly{75.1} & \scoreonly{61.9} & \scoreonly{72.0} & \scoreonly{65.2} & \scoreonly{24.5} & \scoreonly{59.7} \\

Dream-Coder* & Qwen2.5-Coder-7B 
& \scoreonly{75.9} & \scoreonly{61.6} & \scoreonly{66.5} & \scoreonly{60.4} & \scoreonly{21.4} & \scoreonly{57.2} \\

d1* & LLaDA+RL 
& \scoreonly{39.0} & \scoreonly{--} & \scoreonly{45.5} & \scoreonly{--} & \scoreonly{--} & \scoreonly{--} \\

\midrule
\rowcolor{subsectionbg}
\multicolumn{8}{c}{Looped Latent Reasoning Models} \\

Ouro* & from scratch (2.6B) 
& \scoreonly{80.4} & \scoreonly{66.6} & \scoreonly{78.2} & \scoreonly{70.7} & \scoreonly{\textbf{38.7}} & \scoreonly{66.9} \\

\midrule
\rowcolor{sectionbg}
\multicolumn{8}{c}{\textbf{Method Comparison}} \\
\rowcolor{subsectionbg}
\multicolumn{8}{c}{Non-RL Methods} \\

Base Model & Qwen3-8B-Base 
& \scoreonly{60.5} 
& \scoreonly{53.8} 
& \scoreonly{78.2} 
& \scoreonly{68.6} 
& \scoreonly{17.7}
& \scoreonly{55.8} \\

Standard SFT & Qwen3-8B-Base 
& \scorecell{63.3}{\pos{2.8}} 
& \scorecell{52.7}{\decr{1.1}} 
& \scorecell{84.6}{\pos{6.4}} 
& \scorecell{69.5}{\pos{0.9}} 
& \scoreonly{--}
& \scoreonly{--} \\

\rowcolor{latentbg}
Soft Thinking & Qwen3-8B-Base 
& \scorecell{64.2}{\pos{3.7}} 
& \scorecell{53.1}{\decr{0.7}} 
& \scorecell{85.0}{\pos{6.8}} 
& \scorecell{71.2}{\pos{2.6}} 
& \scoreonly{--}
& \scoreonly{--} \\

\rowcolor{latentbg}
TaH+ & Qwen3-8B-Base 
& \scorecell{65.6}{\pos{5.1}} 
& \scorecell{56.5}{\pos{2.7}} 
& \scorecell{\underline{85.8}}{\pos{7.6}} 
& \scorecell{74.3}{\pos{5.7}} 
& \scoreonly{--}
& \scoreonly{--} \\

\rowcolor{latentbg}
LaVAE & Qwen3-8B-Base 
& \scorecell{42.0}{\decr{18.5}} 
& \scorecell{30.2}{\decr{23.6}} 
& \scorecell{47.8}{\decr{30.4}} 
& \scorecell{32.8}{\decr{35.8}} 
& \scorecell{10.8}{\decr{6.9}}
& \scorecell{32.7}{\decr{23.0}} \\

\rowcolor{latentbg}
LaDiR & Qwen3-8B-Base 
& \scorecell{66.8}{\pos{6.3}} 
& \scorecell{59.5}{\pos{5.7}} 
& \scorecell{\textbf{87.4}}{\pos{9.2}} 
& \scorecell{73.2}{\pos{4.6}} 
& \scorecell{21.3}{\pos{3.6}}
& \scorecell{61.6}{\pos{5.9}} \\

\rowcolor{latentbg}
\textbf{\methodname{} (Dual-Path)} & Qwen3-8B-Base
& \scorecell{77.5}{\pos{17.0}} 
& \scorecell{66.2}{\pos{12.4}} 
& \scorecell{82.9}{\pos{4.7}} 
& \scorecell{\underline{77.7}}{\pos{9.1}} 
& \scorecell{21.6}{\pos{3.9}}
& \scorecell{65.2}{\pos{9.4}} \\

\rowcolor{latentbg}
\textbf{\methodname{} (Unified)} & Qwen3-8B-Base
& \scorecell{\textbf{83.9}}{\pos{23.4}} 
& \scorecell{\textbf{72.1}}{\pos{18.3}} 
& \scorecell{\underline{85.8}}{\pos{7.6}} 
& \scorecell{\textbf{78.3}}{\pos{9.7}} 
& \scorecell{23.7}{\pos{6.0}}
& \scorecell{\textbf{68.8}}{\pos{13.0}} \\

\rowcolor{latentbg}
\textcolor{gray}{\quad$\hookrightarrow$ RL} & \textcolor{gray}{Qwen3-8B-Base}
& \textcolor{gray}{\scorecell{85.4}{\pos{24.9}}}
& \textcolor{gray}{\scorecell{73.3}{\pos{19.5}}}
& \textcolor{gray}{\scorecell{86.7}{\pos{8.5}}}
& \textcolor{gray}{\scorecell{80.2}{\pos{11.6}}}
& \textcolor{gray}{\scorecell{25.1}{\pos{7.4}}}
& \textcolor{gray}{\scorecell{70.1}{\pos{14.3}}} \\

\bottomrule
\end{tabular}%
}

\label{tab:main_results}
\end{table*}
\subsection{Experimental Setup}

\paragraph{Datasets and evaluation.}
We fine-tune \methodname{} on Ling-Coder~\citep{codefuse2025samplemattersleveragingmixtureofexperts},
a dataset of 1.4M Python instruction-following samples. We evaluate on five
Python code benchmarks: HumanEval~\citep{chen2021evaluating},
MBPP~\citep{austin2021program}, their EvalPlus versions HumanEval+ and
MBPP+~\citep{liu2023is}, and LiveCodeBench v6~\citep{jain2024livecodebench}.
HumanEval+ and MBPP+ add more test cases for stricter evaluation. For each
problem, we sample 16 solutions and report the average pass@1 to reduce
variance. For the RL stage, we further train on data
from AceCoder~\citep{zeng2025acecoderacingcoderrl} and KodCode~\citep{xu2025kodcodediversechallengingverifiable}.

\paragraph{Baselines.}
We compare \methodname{} with strong open-source models and latent reasoning
baselines. Autoregressive baselines include
Qwen2.5-Coder~\citep{hui2024qwen25coder}, OpenCoder~\citep{huang2024opencoder},
OlympicCoder~\citep{penedo2025olympiccoder}, and
Seed-Coder~\citep{seed2025seedcoder}; diffusion baselines~\citep{ho2020ddpm}
include Dream~\citep{ye2025dream7b}, LLaDA~\citep{nie2025llada},
Diffu-Coder~\citep{gong2025diffucoder}, Dream-Coder~\citep{xie2025dreamcoder},
and d1~\citep{zhao2025d1}; we also include the looped latent reasoning model
Ouro~\citep{zhu2025ouro}. Under the same Qwen3-8B-Base
backbone~\citep{yang2025qwen3}, we further compare with Standard SFT,
Soft Thinking~\citep{zhang2026soft}, TaH+~\citep{fu2025think},
LaVAE (replaces our flow with a $L_2$ objective VAE) and
LaDiR~\citep{ladir2025} (replaces our flow with latent diffusion). We report
two variants of our method, \methodname{} (Dual-Path) and \methodname{}
(Unified).

\paragraph{Experiment details.}
We use Qwen3-8B-Base with $N = 64$ latent slots and a flow stack of $K = 5$ shallow MetaBlocks (alternating Identity/Flip permutations) followed by the deep LLM block. Full training and inference details are in Appendix~\ref{app:training_details}.

\subsection{Main Results}
\label{subsec:main_results}

\paragraph{Pass@1 improvements on code generation.}
Table~\ref{tab:main_results} reports pass@1 on five code benchmarks. \methodname{} (Unified) achieves the best average performance, improving Qwen3-8B-Base from 55.8 to 68.8 (+13.0\%), while \methodname{} (Dual-Path) also improves the base by +9.4\%. On the four shared non-LCB benchmarks, Standard SFT improves the base from 65.3 to 67.5, while \methodname{} (Unified) reaches 80.0 with the same training data. This suggests that the gains come from modeling the distribution of reasoning trajectories, not simply from more CoT data. Among prior latent reasoning methods, \methodname{} (Unified) outperforms the strongest baseline LaDiR with a +7.1\% gain on average. This shows that learning a continuous CoT distribution with normalizing flow captures latent reasoning structure more effectively. \methodname{} (Unified) also surpasses the strongest open-source baselines OlympicCoder (+0.3\%), Diffu-Coder (+9.1\%), and the looped latent reasoning model Ouro (+1.9\%) in average pass@1.

\paragraph{Pass@$k$ scaling}

We test whether latent reasoning scales with more samples by comparing
\methodname{} with the Qwen3-8B base model and LaDiR on MBPP+ and HumanEval+
for $k$ up to 128.

\begin{figure}[t]
    \centering
    \includegraphics[width=0.95\linewidth]{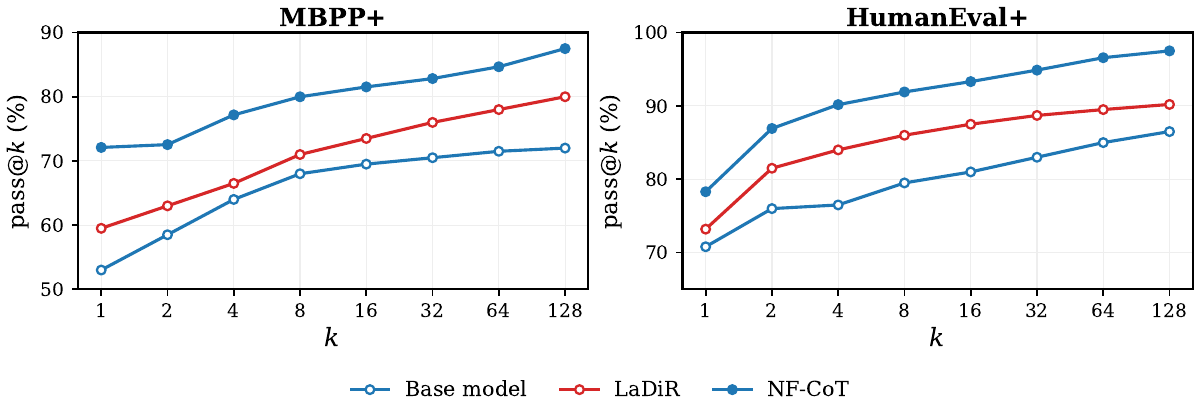}
    \caption{
    Pass@$k$ scaling on MBPP+ and HumanEval+. \methodname{} outperforms the
    base model and LaDiR, and continues to improve with larger $k$.
    }
    \label{fig:passk}
\end{figure}

Figure~\ref{fig:passk} shows that \methodname{} dominates both baselines
across the full $k$ range in this separate scaling run. On MBPP+,
\methodname{}'s pass@1 ($72.1$) already matches the base model's pass@128
($72.0$) and rises to $87.5$ at $k=128$. On HumanEval+, \methodname{} improves
from $78.3$ to $97.5$ (+19.2), comparable to the base model's +16.5 despite
starting from a much higher pass@1. LaDiR, in contrast, only improves from
$73.2$ to $90.2$ (+17.0), with the curve flattening at larger budgets. This
suggests that sampling the flow noise produces distinct continuous-CoT
trajectories rather than collapsing onto a single solution mode. We further
verify this with pairwise structural similarity among passing programs in
Appendix~\ref{app:pairwise_diversity}, confirming structural diversity.
Qualitative decoded-latent examples in
Appendix~\ref{app:latent_cot_showcase} provide example-level evidence for the
same phenomenon: different latent samples for the same HumanEval prompt can
steer decoding toward distinct correct implementation strategies. We treat
these decoded latent CoTs only as qualitative probes, not as faithful
natural-language explanations of the model's internal reasoning process.

\paragraph{Execution-guided latent RL.}
We further apply GRPO to the supervised \methodname{} (Unified) using execution rewards on a random selected 20K-problem Python mixture for 150
training steps. As shown in Table~\ref{tab:main_results}, this lightweight RL
stage consistently improves \methodname{} across all five benchmarks, raising
the average from $68.8$ to $70.1$. These modest gains show that the tractable
likelihood of \methodname{} provides a direct interface for policy
optimization in latent space, allowing execution rewards to update not only
the answer decoder but also the distribution over latent reasoning paths.
RL setup details are in Appendix~\ref{app:rl_details}.

\begin{figure}[t]
\centering
\includegraphics[width=0.95\linewidth]{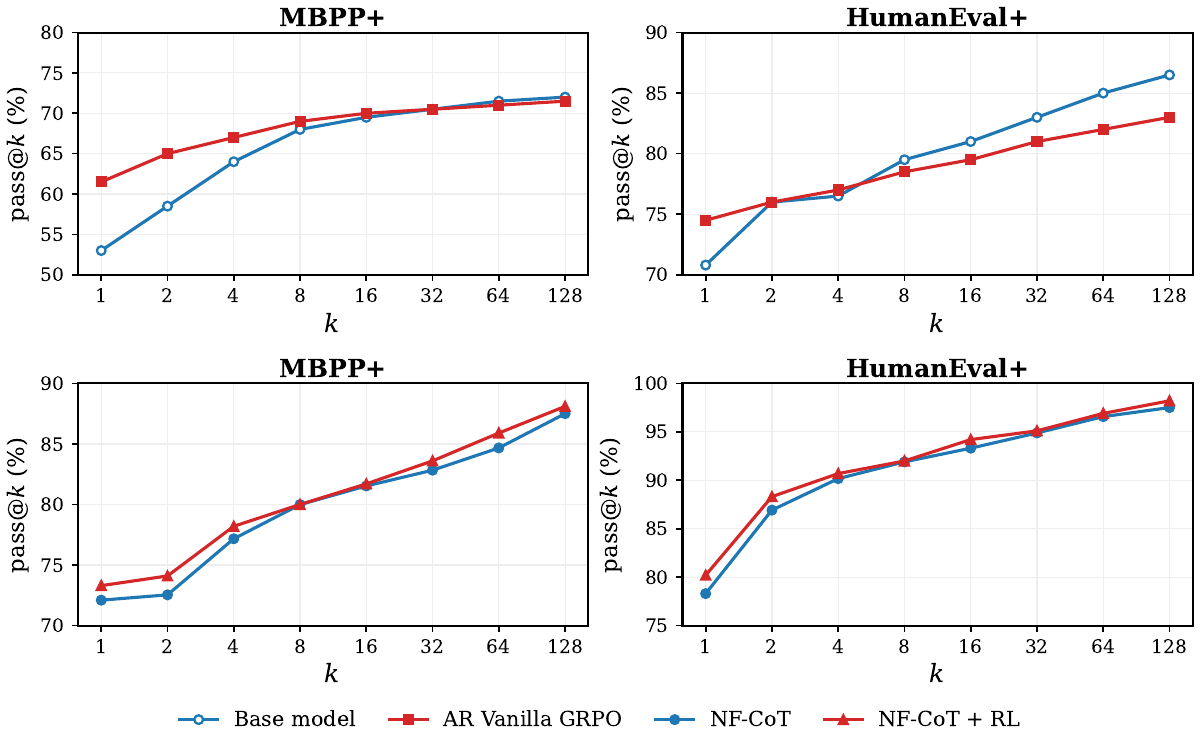}
\caption{
Pass@k diversity before and after reinforcement learning on MBPP+ and HumanEval+.
Top row: standard token-space GRPO improves the low-$k$ region but saturates at larger $k$,
failing to provide the same large-sample coverage as the base model. Bottom row: latent-space
RL for NF-CoT improves pass@1 while preserving the upward pass@k scaling trend, indicating
that policy-gradient refinement in continuous-CoT space does not collapse the latent trajectory
distribution.
}
\label{fig:rl_passk_diversity}
\end{figure}

\paragraph{RL preserves pass@k diversity.}
A common concern for reinforcement learning on autoregressive language models is that optimizing a
single correctness reward can increase pass@1 while reducing the diversity needed for pass@k scaling.
We therefore compare standard token-space GRPO with our latent-space RL under the same pass@k
diagnostic on MBPP+ and HumanEval+. As shown in Figure~\ref{fig:rl_passk_diversity}, the two
RL settings exhibit different large-$k$ behavior. In the token-space setting, AR Vanilla GRPO improves
the low-$k$ region but saturates below, or only matches, the base model at large $k$. This indicates that
token-space RL can concentrate probability mass on a smaller set of solution modes: the model becomes
more likely to find one correct solution, but additional samples do not provide substantially broader
coverage.

In contrast, applying RL to NF-CoT does not produce this pass@k collapse. NF-CoT + RL improves
pass@1 while preserving the upward scaling trend at larger sampling budgets, and it remains consistently
above the supervised NF-CoT checkpoint across the full $k$ range on both benchmarks. This suggests that
policy-gradient refinement in the learned continuous-CoT space improves solution quality without collapsing
the latent trajectory distribution. Together with the exact latent likelihood in Eq.~(3.4), this supports the
role of NF-CoT as an RL-compatible latent reasoning policy: execution rewards can refine both answer
decoding and latent trajectory sampling, while maintaining the diversity required for pass@k gains.

\section{Analysis}
\label{sec:analysis}

\begin{table*}[t]
\centering
\caption{HumanEval inference efficiency with vLLM answer decoding. 
All methods generate $16$ candidates per problem; LaDiR uses $30$ denoising steps.
``Latent'' and ``Decode'' report latent-generation and answer-decoding time, respectively.
FLOPs/sample are coarse estimates from saved outputs.}
\footnotesize
\setlength{\tabcolsep}{4pt}
\begin{tabular}{l c c c c c}
\toprule
Method & Latent (s) $\downarrow$ & Decode (s) $\downarrow$ & Total (s) $\downarrow$ & Samples/s $\uparrow$ & FLOPs/sample $\downarrow$ \\
\midrule
\methodname{} (Unified) & \textbf{173.5} & 152.1 & \textbf{325.6} & \textbf{8.06} & \textbf{19.9T} \\
\methodname{} (Dual-Path) & 232.3 & \textbf{147.7} & 380.0 & 6.90 & 21.6T \\
LaDiR & 468.2 & 157.1 & 625.3 & 4.20 & 49.3T \\
\bottomrule
\end{tabular}
\label{tab:inference_efficiency_vllm}
\end{table*}

\begin{table}[t]
\centering
\caption{Stage-2 training throughput and compute. Note that LaDiR uses its paper setting of $30$ denoising steps.}
\footnotesize
\setlength{\tabcolsep}{6pt}
\begin{tabular}{l c c c}
\toprule
Method & Samples/s $\uparrow$ & Tokens/s $\uparrow$ & Total FLOPs $\downarrow$ \\
\midrule
\methodname{} (Unified) & \textbf{18.4} & \textbf{5.88K} & \textbf{2.25e19} \\
\methodname{} (Dual-Path) & 12.2 & 3.91K & \textbf{2.25e19} \\
LaDiR & 6.45 & 1.03K & 1.50e20 \\
\bottomrule
\end{tabular}
\label{tab:training_efficiency}
\end{table}

\subsection{Training and inference efficiency}
\label{subsec:efficiency}
Tables~\ref{tab:inference_efficiency_vllm} and~\ref{tab:training_efficiency}
compare \methodname{} with LaDiR on HumanEval under matched settings
(16 candidates per problem, Qwen3-8B backbone, vLLM decoding). Both methods
represent reasoning using only 64 latent tokens, which on average encode a
short CoT of 385 text tokens, corresponding to an effective compression rate
of approximately $6.0\times$ compared to the explicit-CoT method. The key
difference is in latent generation: LaDiR runs 30 denoising steps over
the latent trajectory, while \methodname{} samples 64 continuous thoughts
autoregressively in one left-to-right pass. This makes \methodname{}
(Unified) $2.70\times$ faster at latent generation ($468.2 \to 173.5$s),
$1.92\times$ faster overall ($625.3 \to 325.6$s), and $2.48\times$ cheaper
in per-sample compute ($49.3 \to 19.9$T FLOPs). The dual-path variant is
slower than Unified because it keeps separate objective-specific paths,
but still $1.65\times$ faster and $2.29\times$ cheaper than LaDiR. The
same trend holds during training: \methodname{} (Unified) achieves
$2.85\times$ sample and $5.71\times$ token throughput over LaDiR, with
$6.66\times$ fewer total FLOPs; \methodname{} (Dual-Path) also reaches
$1.90\times$ higher sample throughput. These results support our
efficiency claim: replacing
iterative denoising with an autoregressive flow speeds up latent generation,
while exact-likelihood training avoids learning a multi-step denoising chain
required by diffusion-based methods such as LaDiR.



\begin{table}[t]
\centering
\caption{
Ablation of the frozen-backbone warm-up. Removing Stage~1 hurts three benchmarks (pass@1).
}
\footnotesize
\setlength{\tabcolsep}{3pt}
\begin{tabular}{l c c c}
\toprule
Benchmark & Stage 1+2 & Stage-2-only & $\Delta$ \\
\midrule
HumanEval & \textbf{84.4} & 81.5 & $-2.9$ \\
HumanEval+ & \textbf{78.7} & 75.5 & $-3.2$ \\
LiveCodeBench v6 & \textbf{23.1} & 21.4 & $-1.7$ \\
\bottomrule
\end{tabular}
\label{tab:stage1_ablation}
\end{table}

\subsection{Ablation study}

\paragraph{Effect of the frozen-backbone warm-up stage.}
We ablate the two-stage training curriculum from
Section~\ref{sec:method}. In the default setting, \methodname{} first
trains the flow components with the LLM backbone frozen, then unfreezes
all parameters for joint optimization. We compare this with a
stage-2-only variant that skips the warm-up. As shown in
Table~\ref{tab:stage1_ablation}, removing Stage~1 consistently hurts
performance: HumanEval drops from 84.4 to 81.5, HumanEval+ from 78.7 to
75.5, and LiveCodeBench v6 from 23.1 to 21.4. This is not due to worse
final optimization: both runs reach nearly the same total loss and
answer cross-entropy. The difference appears in the early training
dynamics. With the warm-up, the flow components already provide a
meaningful latent interface before the backbone is updated; without it,
the backbone receives early gradients from randomly initialized flow
components, which can disrupt the coding capability we aim to preserve.
The warm-up thus acts as a curriculum that aligns the latent-flow
interface with a frozen backbone before allowing the NF likelihood
objective to update the full model. Appendix~\ref{app:backbone_drift}
provides additional backbone-drift diagnostics.

\paragraph{Unified versus dual-path training.}
We next ablate the unified causal training design. The dual-path variant
uses separate paths for the NF and CE objectives: one path trains the NF
head on continuous thoughts, the other trains the LM head to decode
answers from a different latent interface. Although the two paths share
the backbone, they use different prefix tokens, projections, and
parameterizations, creating a mismatch between the latent distribution
learned by NF, the latent representation used for answer conditioning, and
the trajectory used at inference. The unified variant removes this
mismatch by placing latent reasoning and answer decoding in a single
causal sequence: the same continuous-CoT prefix is scored by the NF head
and then used by the LM head for answer prediction, so the NF and CE
losses are defined on the same context and the same trajectory used at
inference. This structural alignment explains why \methodname{} (Unified)
outperforms the dual-path variant: the gain comes from removing redundant
latent interfaces and training continuous-CoT states to be useful for
answer decoding, avoiding a second backbone pass and prefix recomputation.

\subsection{Latent Perturbation Robustness}
\label{sec:latent_perturbation}

\begin{figure*}[t] \centering \includegraphics[width=\linewidth,trim=3pt 3pt 3pt 3pt,clip]{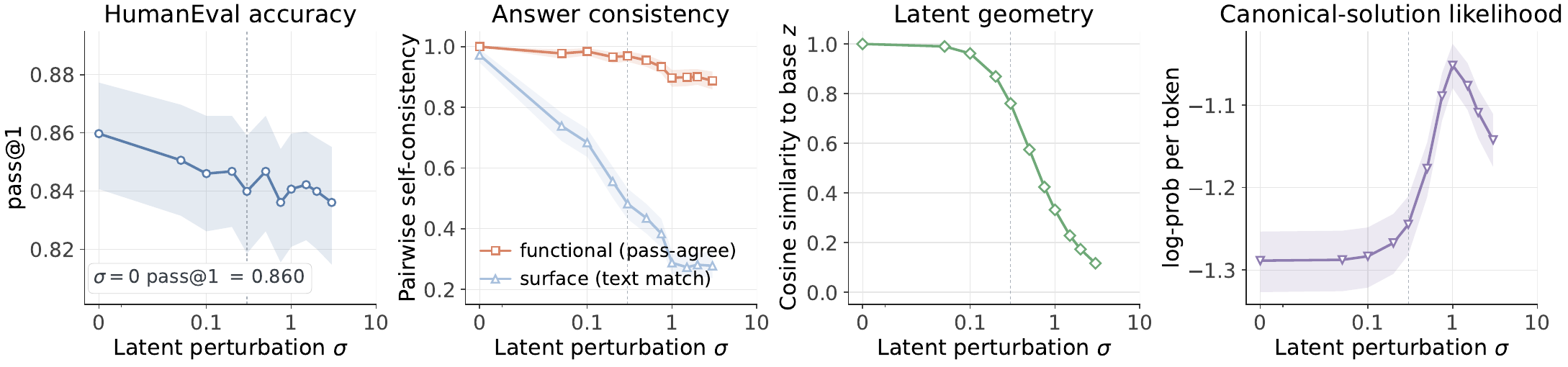} \caption{Effect of continuous-CoT perturbation strength on HumanEval generation. As $\sigma$ increases, the perturbed continuous-CoT trajectory becomes nearly orthogonal to the original trajectory, and exact-text match drops sharply. Nevertheless, pass@1 remains nearly flat, indicating that local perturbations change the form of the generated program much more than its functional correctness, even at the largest perturbation strengths tested.} \label{fig:latent_perturbation} \end{figure*}

We test whether \methodname{} learns a smooth latent reasoning space. For
each HumanEval problem, we sample a base continuous-CoT trajectory
$\tilde{u}$, perturb it with Gaussian noise
\begin{equation}
u_{\sigma} = \tilde{u} + \sigma \epsilon,
\qquad \epsilon \sim \mathcal{N}(0, I),
\end{equation}
and greedily decode from the perturbed prefix, measuring pass@1, pairwise
pass agreement, exact-text match, cosine similarity, and the
canonical-solution log-probability. As $\sigma$ grows from $0$ to $3.0$
(Figure~\ref{fig:latent_perturbation}), cosine similarity drops from
$1.000$ to $0.116$, but pass@1 only decreases from $86.0$ to $83.6$
with overlapping confidence intervals, indicating that the learned
continuous-CoT space is locally smooth. Perturbations affect form much
more than function: pairwise pass agreement drops only slightly
($1.000 \to 0.888$), while exact-text match drops sharply
($0.973 \to 0.278$). Continuous-CoT thus acts less like brittle hidden
tokens and more like distributed control variables over solution
trajectories, perturbations switch implementation style rather than
break correctness. The canonical-solution log-probability follows a
non-monotonic trend: moderate noise raises it (the original trajectory
commits to one valid solution mode that may differ from the canonical
implementation), while large noise degrades the prefix. Overall,
\methodname{}'s latent variables control which solution the model follows,
not whether it solves the problem.

\section{Related Work}
\label{sec:related}

\noindent\textbf{CoT and latent reasoning.}
Chain-of-thought (CoT) prompting elicits reasoning by generating
intermediate steps before the answer~\citep{wei2022chain,kojima2022large},
with further gains from supervised or RL-based
training~\citep{zelikman2022star,shao2024deepseekmath} and test-time
strategies that sample and score many
chains~\citep{wang2022self,snell2024scaling,yao2023tot}. These approaches
all reason in token space. Latent reasoning instead feeds continuous
thoughts back into the
model~\citep{hao2024coconut,zhang2026soft,tang2026multiplex,su2025tokenassorted},
but probing shows each collapses to a single greedy path, not a
superposition~\citep{wu2026singlethreaded}. Recent
work injects stochasticity to enable RL or model trajectory superpositions,
including in the multimodal
setting~\citep{butt2025softtokens,zhou2026lepo,deng2026superposition,wang2025monet}, several
via Gumbel-Softmax relaxations that turn token choices into soft embedding
mixtures~\citep{jang2017categorical,maddison2017concrete}, but none place a tractable density over the trajectories. Others run
multiple latent
channels or internal recurrences without modeling a trajectory
distribution~\citep{geiping2025recurrent,zhu2025ouro,fu2025think}. The closest
precedent denoises a VAE latent space, leaving the trajectory
density implicit and the log-likelihood
intractable~\citep{ladir2025,ladirl2026}. \textsc{\methodname{}} instead
uses an autoregressive normalizing flow over VAE latents inside the LLM's
causal stream, giving an exact log-likelihood over continuous-CoT
trajectories and training latent reasoning and answer generation jointly.

\noindent\textbf{Normalizing flows.}
Normalizing flows map a simple base distribution to data through
invertible networks, and provide
exact, tractable
likelihoods~\citep{rezende2015variational}. Early work
applied them to images, showing that flows can scale
to natural image
generation~\citep{dinh2014nice,dinh2016density,kingma2018glow}. A
parallel line adapted flows to NLP by lifting discrete
text into continuous latent spaces,
with applications to language modeling and
non-autoregressive
translation~\citep{ziegler2019latent,ma2019flowseq}.
Autoregressive flows~\citep{kingma2016improved,papamakarios2017maf},
whose triangular Jacobian aligns naturally with causal Transformers,
have more recently been scaled to high-resolution images and
video~\citep{zhai2024tarflow,gu2025starflow,gu2025starflowv,gu2026ntm}
and, most recently, to text, either by unifying image
and text generation or by modeling language as a
continuous latent
sequence~\citep{shen2026starflow2,zhang2026flexible}.
In contrast, \textsc{\methodname{}} uses autoregressive
flows inside a pretrained LLM as a tool for
\emph{reasoning}, modeling a distribution over continuous CoT trajectories.









\section{Conclusion}

We presented \methodname{}, a latent reasoning framework that gives
continuous CoT the same modeling status as language tokens by running an
autoregressive normalizing flow inside the LLM's causal stream.
\methodname{} models an explicit distribution over reasoning
trajectories with exact likelihood, which supports both supervised
likelihood training and policy-gradient refinement in the continuous
reasoning space. Across code generation benchmarks, \methodname{}
improves accuracy over SFT and prior latent-reasoning baselines on
Qwen3-8B-Base, and runs faster than LaDiR in training and inference.
These results suggest that likelihood-based latent reasoning offers a
practical interface for sampling, scoring, and refining continuous
thoughts in LLMs.

\section*{Limitations}

Our validation focuses on code-generation benchmarks; extending to other
reasoning tasks remains future work. \methodname{} is trained on fixed-length
VAE-encoded trajectories from explicit-CoT traces, so the learned density
inherits their biases, and a fixed latent budget may not fit all problems.
Latent likelihood is not a correctness oracle, and our RL refinement relies on
unit-test rewards available only in code; generalization to domains without
verifiers is open. Improved code generation may also produce plausible-yet-incorrect
or unsafe code, and continuous latents are not human-readable, so decoded CoTs
are qualitative probes rather than faithful explanations of internal reasoning.

\bibliographystyle{plainnat}
\bibliography{custom}

\clearpage
\appendix
\section{Qualitative Analysis of Decoded Latent CoTs}
\label{app:latent_cot_showcase}

In this appendix, we provide qualitative examples of decoded latent CoTs on HumanEval. Each example contains the raw text decoded from a sampled latent CoT vector and the program generated from the same latent prefix. The decoded latent text is not edited for correctness or fluency. Because this text is reconstructed through the frozen VAE decoder, we do not interpret it as a faithful natural-language explanation of the model's internal computation. Instead, we use it as a qualitative probe of whether sampled latent trajectories align with coarse algorithmic choices in the final program. These examples show that \methodname{}'s latent variables often correspond to high-level algorithmic choices: different latent samples for the same problem can lead to different implementation strategies while still producing functionally correct programs.

\paragraph{Experimental setup.}
We use the \methodname{} unified checkpoint with 64 latent tokens. For each task, we sample latent trajectories with guidance scale $1.0$, NF temperature $0.9$, and answer temperature $0.6$. The decoded latent CoT is obtained by converting the unified latent back to the VAE-latent space and greedily decoding it with the VAE decoder. The benchmark subset is from HumanEval and is not part of the training data.

\begin{table}[h]
\centering
\small
\caption{Pass counts for the HumanEval tasks considered in the latent-CoT
showcase. We provide detailed decoded examples for a representative subset
below.}
\label{tab:latent_cot_showcase_counts}
\begin{tabular}{l l c}
\toprule
Task & Entry point & Passed / 64 \\
\midrule
HumanEval/75 & \texttt{is\_multiply\_prime} & 55 / 64 \\
HumanEval/124 & \texttt{valid\_date} & 64 / 64 \\
HumanEval/154 & \texttt{cycpattern\_check} & 61 / 64 \\
HumanEval/63 & \texttt{fibfib} & 64 / 64 \\
HumanEval/44 & \texttt{change\_base} & 64 / 64 \\
HumanEval/38 & \texttt{decode\_cyclic} & 64 / 64 \\
HumanEval/123 & \texttt{get\_odd\_collatz} & 58 / 64 \\
HumanEval/89 & \texttt{encrypt} & 64 / 64 \\
\bottomrule
\end{tabular}
\end{table}

\subsection{HumanEval/63: \texttt{fibfib}}

\begin{cocotaskbox}{Problem}
\begin{lstlisting}[style=cocopystyle, language=Python]
def fibfib(n: int):
    """The FibFib number sequence is a sequence similar to the Fibbonacci sequnece that's defined as follows:
    fibfib(0) == 0
    fibfib(1) == 0
    fibfib(2) == 1
    fibfib(n) == fibfib(n-1) + fibfib(n-2) + fibfib(n-3).
    Please write a function to efficiently compute the n-th element of the fibfib number sequence.
    >>> fibfib(1)
    0
    >>> fibfib(5)
    4
    >>> fibfib(8)
    24
    """
\end{lstlisting}
\end{cocotaskbox}

This task is a strong showcase because the same problem produces three passing answers with genuinely different implementation strategies: rolling-state dynamic programming, explicit tabulation, and recursive memoization.

\begin{latentbox}{sample\_idx=49, rolling three-state DP}
\textit{Why useful:} The decoded latent CoT discusses compressing the recurrence into fixed state, and the generated answer uses only \texttt{a}, \texttt{b}, and \texttt{c}.

\begin{lstlisting}[style=cocopystyle]
To calculate the Fibonacci number sequence up to the \( n \)th number, we need to follow these steps:

### Plan

1. **Understanding the Fibonacci Sequence**: The Fibonacci sequence is defined as:
   - \( F(1) = 0 \)
   - \( F(2) = 1 \)
   - \( F(n) = F(n-1) + F(n-2) \)
2. **Compressing the Formula for Efficiency**: We use a fixed amount of space to store the last three Fibonacci numbers and calculate the next. This optimizes, since we don't need to use \( n \) variables.
3. **Recursive Fibonacci Term**: Calculate the \( n \)th Fibonacci number:
   - \( F(n) = F(n-1) + F(n-2) \)
4. **Flow**: This calculation is done in constant \( O(1) \) extra space by using three variables.
\end{lstlisting}
\end{latentbox}

\begin{answerbox}{sample\_idx=49, passed}
\begin{lstlisting}[style=cocopystyle, language=Python]
def fibfib(n: int) -> int:
    """Compute the n-th element of the fibfib sequence."""
    # Base cases
    if n == 0:
        return 0
    elif n == 1:
        return 0
    elif n == 2:
        return 1

    # Initialize the first three elements of the sequence
    a, b, c = 0, 0, 1

    # Compute the sequence iteratively up to the n-th element
    for _ in range(3, n + 1):
        # Calculate the next element in the sequence
        next_value = a + b + c
        # Update the previous three elements for the next iteration
        a, b, c = b, c, next_value

    return c
\end{lstlisting}
\end{answerbox}

\begin{latentbox}{sample\_idx=3, explicit tabulation list}
\textit{Why useful:} The decoded latent CoT discusses tabulation/list storage, and the generated answer appends every FibFib value to \texttt{fibfib\_values}.

\begin{lstlisting}[style=cocopystyle]
To solve the Fibonacci number sequence using a tabular approach, we can efficiently compute the nth Fibonacci number and compute the sequence for the first n Fibonacci numbers. The Fibonacci number is defined by the formula:

- `fib(n) = fib(n-1) + fib(n-2)`

Here's a step-by-step breakdown of how we can implement this:

### Step-by-Step Explanation

1. **Initialization:**
   - We use a stack (list) to store Fibonacci numbers, starting with the base cases `n = 0` and `n = 1`.
   - This prevents the exponential number of computations and ensures the sequence is computed efficiently.
2. **Base Case:**
   - The first two Fibonacci numbers are known: `fib(0) = 0` and `fib(1) = 1`.
3. **Compute the Sequence:**
   - For a given `n`, compute the value of `fib(n)` by summing the previous two Fibonacci numbers (last and nth).
   - Compute the rest of the sequence numbers: `fib(n-1)`, `fib(n-2)`, etc., from 3 to `n`.
4. **Return Computed Values:**
   - Return the array of computed Fibonacci numbers.
5. **Implementation**: The function computes the necessary Fibonacci number for `n`, and indeed utilizes efficient tabulation to avoid redundant computations.
\end{lstlisting}
\end{latentbox}

\begin{answerbox}{sample\_idx=3, passed}
\begin{lstlisting}[style=cocopystyle, language=Python]
def fibfib(n: int) -> int:
    """Compute the n-th element of the FibFib sequence."""
    # Base cases
    if n == 0:
        return 0
    if n == 1:
        return 0
    if n == 2:
        return 1

    # Initialize the first three values of the sequence
    fibfib_values = [0, 0, 1]

    # Compute the sequence iteratively up to n
    for i in range(3, n + 1):
        # Calculate the next value as the sum of the last three values
        next_value = fibfib_values[-1] + fibfib_values[-2] + fibfib_values[-3]
        # Append the new value to the list
        fibfib_values.append(next_value)

    # Return the n-th value in the sequence
    return fibfib_values[n]
\end{lstlisting}
\end{answerbox}

\begin{latentbox}{sample\_idx=35, recursive memoization}
\textit{Why useful:} The decoded latent CoT discusses cache and recursion, and the generated answer uses a nested memoized helper.

\begin{lstlisting}[style=cocopystyle]
To solve this problem, we can use a cache to efficiently compute the nth Fibonacci number. We'll use a recursive function to compute the Fibonacci numbers, but we'll optimize the recursion by storing the results of previous Fibonacci numbers. This method avoids the exponential number of calculations and efficiently fills the cache, leading to a significant efficiency improvement. Here's the detailed version:

## Efficient Fibonacci Sequence

### Step-by-Step Explanation

1. **Define the Function:**
   - Use a cache (dictionary) to store the value of Fibonacci numbers we have computed.
2. **Recursive Computation:**
   - We compute the Fibonacci number by first calculating the function recursively and store it in the cache.
   - Use base cases to reduce complexity.
\end{lstlisting}
\end{latentbox}

\begin{answerbox}{sample\_idx=35, passed}
\begin{lstlisting}[style=cocopystyle, language=Python]
def fibfib(n: int) -> int:
    """Compute the n-th element of the FibFib sequence."""

    # Initialize a dictionary to store the computed values
    memo = {0: 0, 1: 0, 2: 1}

    def helper(x):
        """Recursive helper function with memoization."""
        # Check if the value is already computed
        if x in memo:
            return memo[x]

        # Compute the value recursively and store it in the memo dictionary
        memo[x] = helper(x - 1) + helper(x - 2) + helper(x - 3)
        return memo[x]

    # Call the helper function with the input n
    return helper(n)
\end{lstlisting}
\end{answerbox}

\subsection{HumanEval/154: \texttt{cycpattern\_check}}

\begin{cocotaskbox}{Problem}
\begin{lstlisting}[style=cocopystyle, language=Python]
def cycpattern_check(a , b):
    """You are given 2 words. You need to return True if the second word or any of its rotations is a substring in the first word
    cycpattern_check("abcd","abd") => False
    cycpattern_check("hello","ell") => True
    cycpattern_check("whassup","psus") => False
    cycpattern_check("abab","baa") => True
    cycpattern_check("efef","eeff") => False
    cycpattern_check("himenss","simen") => True

    """
\end{lstlisting}
\end{cocotaskbox}

This example shows that different latent samples can instantiate the same rotation-substring idea with different surface forms.

\begin{latentbox}{sample\_idx=0, direct rotation loop}
\textit{Why useful:} The decoded latent CoT says to inspect all rotations, and the generated answer loops over rotations with early return.

\begin{lstlisting}[style=cocopystyle]
## solve the problem, we need to check for a pattern in a string rotated a certain number of times. The first step is to rotate the string and the second step is to inspect all rotations for the given pattern.

## Step-by-Step Explanation

1. **Understanding Rotations**:
   - A rotation of a string is a substring of the original string.
   - For example, if `s = "abcdef"`, then the rotations are: `a., b., c., d., e., f.` and `b., c., d., e., f., a.`.
2. **Checking Rotations**:
   - We can generate all possible rotations of a string.
   - For each rotation, we check if it matches the pattern.
3. **Rotating and Sorting**:
   - This approach can be somewhat complex for all rotations.
\end{lstlisting}
\end{latentbox}

\begin{answerbox}{sample\_idx=0, passed}
\begin{lstlisting}[style=cocopystyle, language=Python]
def cycpattern_check(a, b):
    """Check if any rotation of the second word is a substring in the first word."""
    # Get the length of the second word.
    len_b = len(b)

    # Check each rotation of the second word.
    for i in range(len_b):
        # Create a rotation by slicing the string.
        rotation = b[i:] + b[:i]

        # Check if this rotation is a substring in the first word.
        if rotation in a:
            return True

    # If no rotation is a substring, return False.
    return False
\end{lstlisting}
\end{answerbox}

\begin{latentbox}{sample\_idx=16, precompute rotations plus \texttt{any}}
\textit{Why useful:} The decoded latent CoT says to generate cyclic permutations, and the generated answer materializes rotations and calls \texttt{any}.

\begin{lstlisting}[style=cocopystyle]
To solve the problem of checking if a word is a substring of any rotation, we need to consider the given function and introduce some checks to check the rotations. Here's how we can implement the function:

### Step-by-Step Explanation

1. **Check the String Rotation:**
   - For each rotation, check if the `check_string` is a substring. Rotations can be generated by checking if the string is a substring of any of its cyclic permutations.
2. **Understanding Rotations:**
   - **Rotations** are permutations of the string. For any given rotation, if you concatenate the string with itself, the original string becomes a substring.
   - Example: If `s = "abcd"`, the rotations are `["abcd", "bcda", "cdab", "dabc"]`.
3. **Check Substring Check:**
   - If `check_string` is a substring of any rotation, it can be checked if it is a substring of `s + s` (the concatenated version of `s`).
4. **Check for No Rotation:**
   - We need to check if the `check_string` is also a substring of the first rotation. A simple check would be:
     - Check if `check_string` is a substring of `a`, `b`, `c`, or `d` for each rotation.
5. **Implement a Check for All Rotations:**
   - Check all rotations by rotating the string and checking the substring condition.
   - If any rotation contains `check_string`, return `True`.
\end{lstlisting}
\end{latentbox}

\begin{answerbox}{sample\_idx=16, passed}
\begin{lstlisting}[style=cocopystyle, language=Python]
def cycpattern_check(a, b):
    if not b:
        return True
    if len(b) > len(a):
        return False

    # Generate all rotations of b
    rotations = [b[i:] + b[:i] for i in range(len(b))]

    # Check if any rotation of b is a substring of a
    return any(rot in a for rot in rotations)
\end{lstlisting}
\end{answerbox}

\subsection{HumanEval/89: \texttt{encrypt}}

\begin{cocotaskbox}{Problem}
\begin{lstlisting}[style=cocopystyle, language=Python]
def encrypt(s):
    """Create a function encrypt that takes a string as an argument and
    returns a string encrypted with the alphabet being rotated. 
    The alphabet should be rotated in a manner such that the letters 
    shift down by two multiplied to two places.
    For example:
    encrypt('hi') returns 'lm'
    encrypt('asdfghjkl') returns 'ewhjklnop'
    encrypt('gf') returns 'kj'
    encrypt('et') returns 'ix'
    """
\end{lstlisting}
\end{cocotaskbox}

This task illustrates another form of latent-controlled implementation diversity: the same Caesar-style cipher is solved either through alphabet-index lookup or ASCII/modulo arithmetic.

\begin{latentbox}{sample\_idx=1, alphabet-index lookup}
\textit{Why useful:} The decoded latent CoT describes alphabet rotation, and the generated answer uses an explicit alphabet string and \texttt{.index()}.

\begin{lstlisting}[style=cocopystyle]
To solve the problem of encrypting a string using a Caesar cipher, we'll implement a function that handles the rotation of characters. Here's a step-by-step explanation of how we can achieve this:

### Step-by-Step Explanation

1. **Handle Case**:
   - Convert the string to handle cases (uppercase and lowercase). This will handle the case where the user strings are partially lowercased.
2. **Handle Edge Cases**:
   - Handle edge cases such as empty strings or the handle of digits properly.
3. **Encoding and Rotation**:
   - Each character in the string should be shifted by the two-digit number provided.
   - If the shift handles the letters beyond 'z' and 'Z', wrap around to the beginning of the alphabet.
4. **Handle Edge Handling**:
   - Repeat the shifted character to handle each character individually.
\end{lstlisting}
\end{latentbox}

\begin{answerbox}{sample\_idx=1, passed}
\begin{lstlisting}[style=cocopystyle, language=Python]
def encrypt(s):
    """Encrypt the string by shifting the alphabet by 4 places."""
    # Define the alphabet
    alphabet = 'abcdefghijklmnopqrstuvwxyz'
    # Calculate the shift amount
    shift = 4
    # Initialize an empty result string
    encrypted = ''

    # Iterate over each character in the input string
    for char in s:
        # Check if the character is a lowercase letter
        if char in alphabet:
            # Find the original position of the character in the alphabet
            original_position = alphabet.index(char)
            # Calculate the new position after shifting
            new_position = (original_position + shift) % 26
            # Append the encrypted character to the result
            encrypted += alphabet[new_position]
        else:
            # If it's not a letter, keep the character as is
            encrypted += char

    return encrypted
\end{lstlisting}
\end{answerbox}

\begin{latentbox}{sample\_idx=23, ASCII modulo arithmetic}
\textit{Why useful:} The decoded latent CoT describes Caesar rotation and wraparound, and the generated answer uses \texttt{ord}/\texttt{chr} modulo arithmetic.

\begin{lstlisting}[style=cocopystyle]
To solve the problem of encrypting a string using a Caesar cipher, we need to follow these steps:

1. **Function Definition**: Create the function `rotate_encrypt` that takes two parameters: the `string` and the `shift`.
2. **Initialize the Result**: Use a variable to shift each character by the specified number. If the character is shifted by 26, it should wrap around the alphabet.
3. **Iterate Through the String**: Iterate through each character in the string and apply the rotation:
   - Shift each letter by the ASCII value of the letter in the opposite alphabet.
   - Use modulo 26 to ensure the shifts are shifted.
4. **Encrypt the Character**: 
   - Initialize the function to add a shifted character to the result.
5. **Return the Encrypted String**: Return the encrypted string.
\end{lstlisting}
\end{latentbox}

\begin{answerbox}{sample\_idx=23, passed}
\begin{lstlisting}[style=cocopystyle, language=Python]
def encrypt(s):
    """Encrypts the input string by rotating the alphabet by 4 positions."""
    encrypted_string = []

    for char in s:
        if 'a' <= char <= 'z':  # Check if the character is a lowercase letter
            # Rotate the character by 4 positions
            new_char = chr((ord(char) - ord('a') + 4) % 26 + ord('a'))
            encrypted_string.append(new_char)
        elif 'A' <= char <= 'Z':  # Check if the character is an uppercase letter
            # Rotate the character by 4 positions
            new_char = chr((ord(char) - ord('A') + 4) % 26 + ord('A'))
            encrypted_string.append(new_char)
        else:
            # Non-alphabet characters are not changed
            encrypted_string.append(char)

    return ''.join(encrypted_string)
\end{lstlisting}
\end{answerbox}

\paragraph{Takeaway.}
Across these examples, decoded latent CoTs are imperfect at the token level but consistently reflect the coarse algorithmic route followed by the final program. The \texttt{fibfib} examples are especially illustrative: the same problem yields rolling-state dynamic programming, explicit table-based dynamic programming, and recursive memoization. This supports the view that \methodname{} does not merely store a single hidden solution trace. Instead, its latent space captures a distribution over plausible reasoning trajectories, allowing different samples to steer generation toward distinct but correct implementations.

\section{Experimental Setup}
\label{sec:appendix-setup}
\label{app:training_details}

We evaluate two variants of our latent-reasoning normalizing flow (NF):
the \emph{dual-path} model, in which the backbone LLM is invoked twice per
training step (once for flow-density estimation and once for cross-entropy
answer prediction), and the \emph{unified-path} model, in which a single
causal forward computes both losses. Both variants share the same backbone,
latent geometry, dataset, two-stage training schedule, and decoding pipeline,
and differ only in the conditioning sequence and which components are
trainable in Stage~1.

\subsection{Dual-Path Model}
\label{ssec:dual-path-setup}

The dual-path model forwards the backbone twice per gradient step.
The two paths use \emph{distinct} latent--to--hidden projectors
(\texttt{flow\_projector} for the flow path, \texttt{vae.latent\_to\_decoder}
for the CE path) and \emph{distinct} boundary tokens (\texttt{BOT} and
\texttt{AE}). Optional diffusion-forcing on the CE path is supported but
disabled in our reported runs.

\subsection{Unified-Path Model}
\label{ssec:unified-path-setup}

The unified model fuses the two paths into a single causal sequence
\[
[\,\mathrm{prompt};\,\texttt{BOT};\,\mathrm{flow\_proj}(\tilde x);\,\mathrm{answer}\,],
\]
where $\tilde x = \pi_{\text{deep}}\bigl(\mathrm{shallow}(x + \epsilon)\bigr)$ is
the latent in post-shallow, post-permutation space and $\epsilon\sim\mathcal{N}(0,\sigma_{\text{dq}}^2 I)$
is dequantization noise (the \emph{same} noise drives NF and CE).
A single backbone forward yields the hidden states used to compute the NF
NLL on positions $[T_p, T_p+N)$ via \texttt{output\_head} and the CE loss on
positions $[T_p+N, T_p+N+T_a)$ via the LM head. Because the shallow
MetaBlocks are invertible, conditioning the CE path on $\tilde x$ is
representationally equivalent to conditioning on raw latents while
eliminating the second backbone forward; the unified model therefore needs
\emph{only one} latent--to--hidden projector and \emph{one} prefix marker.

\subsection{Training Hyperparameters}
\label{ssec:training-hps}

\paragraph{Backbone and latent geometry.}
Both variants use \texttt{Qwen3-8B-Base} as the backbone. The latent
sequence has $N=64$ slots of dimension $D=2560$, identical to the backbone
hidden size. Latents are produced by a frozen transformer-encoder VAE with
encoder \texttt{Qwen3-4B-Base}; the VAE encoder is never updated. Dequantization noise has
$\sigma_{\text{dq}}=0.3$. The flow-density head consists of an MLP
projector ($\mathrm{LayerNorm}\!\to\!\mathrm{Linear}\!\to\!\mathrm{GELU}\!\to\!\mathrm{Linear}$),
five MetaBlocks \cite{gu2025starflow} with channel width 2048, two layers per
block, head dim 64, alternating
$\textsc{Flip}/\textsc{Identity}$ permutations (so the top, deep, block is
$\textsc{Identity}$ and matches the LLM's L2R causal mask), and a zero-init
\texttt{NFOutputHead} that emits affine parameters $(\sigma,\mu)$ with
$\mathrm{softplus}$-positivized scale and $\tanh$ soft-clip at $\pm 4.0$.

\paragraph{Loss.}
The training objective is
\(
\mathcal{L} = \lambda_{\text{NF}}\,\mathcal{L}_{\text{NF}}
            + \lambda_{\text{CE}}\,\mathcal{L}_{\text{CE}}
\)
with $\lambda_{\text{NF}}=\lambda_{\text{CE}}=1$. The flow-NLL has the
standard form
$\mathcal{L}_{\text{NF}}=\tfrac12\mathbb{E}\|z\|_2^2 - \mathbb{E}\sum_k\log\det J_k$,
and the CE term is teacher-forcing cross-entropy on answer tokens with
PAD positions masked out. Classifier-free guidance is supported via a
learned \texttt{null\_embed}; the prompt is dropped to \texttt{null\_embed}
with probability $p_{\text{drop}}=0.1$ (batch-level for the dual-path
model, per-sample for the unified-path model).

\paragraph{Two-stage curriculum.}
We follow a frozen-backbone $\rightarrow$ joint-finetune curriculum:
\begin{itemize}
  \item \textbf{Stage 1 (latent-projector warm-up).} The backbone is frozen.
         The shallow blocks \emph{are} trained in Stage 1.
        Stage 1 uses $100\,000$ samples, 1 epoch, per-device batch size 2,
        gradient accumulation 4 (global batch 32 on 4 GPUs), peak
        learning rate $1\!\times\!10^{-4}$ with 100 linear-warmup steps.
  \item \textbf{Stage 2 (joint finetuning).} The backbone and shallow
        blocks are unfrozen and the full Python subset of the dataset is used for 2 epochs. We use
        per-device batch size 1, gradient accumulation 8 (global batch
        64 on 8 GPUs), peak learning rate $5\!\times\!10^{-5}$, with
        200 linear-warmup steps.
\end{itemize}
For both stages we use AdamW with $(\beta_1,\beta_2)=(0.9, 0.95)$, weight
decay $0.01$, gradient clip $1.0$, cosine learning-rate decay,
\texttt{bfloat16} mixed precision, and FSDP \texttt{hybrid\_shard} with
auto-wrap on \texttt{Qwen3DecoderLayer}, backward/forward prefetching,
CPU-RAM-efficient loading, and activation checkpointing. Training is run
on NVIDIA DGX B200 nodes (4 GPUs in Stage 1, 8 GPUs in Stage 2). The backbone weights for Stage 2
come directly from the pre-trained \texttt{Qwen3-8B-Base} checkpoint.
Training uses the HuggingFace Trainer default seed when no explicit seed is
specified in the configuration. The total number of optimizer steps is
determined by the epoch budgets, global batch sizes, and filtered training
split rather than by a fixed \texttt{max\_steps} override.

\paragraph{Hyperparameter table.}
Table~\ref{tab:hp} summarizes the hyperparameters of both variants.

\begin{table*}[t]
  \centering
  \caption{Hyperparameters for the dual-path and unified-path NF models. Shared settings are shown once across both model columns.}
  \label{tab:hp}
  \scriptsize
  \setlength{\tabcolsep}{4pt}
  \renewcommand{\arraystretch}{1.08}
  \begin{tabular}{p{0.25\textwidth}p{0.34\textwidth}p{0.34\textwidth}}
    \toprule
    \textbf{Setting} & \textbf{Dual-path} & \textbf{Unified-path} \\
    \midrule
    Backbone LLM & \multicolumn{2}{p{0.68\textwidth}}{Qwen3-8B-Base} \\
    Frozen latent encoder & \multicolumn{2}{p{0.68\textwidth}}{TransEncoderVAE with Qwen3-4B-Base encoder} \\
    Latent geometry & \multicolumn{2}{p{0.68\textwidth}}{$N=64$ slots, latent dimension $D=2560$} \\
    Shallow MetaBlocks & 5 blocks, trained in Stage~1 & 5 blocks, trained in Stage~1 \\
    Shallow block width & \multicolumn{2}{p{0.68\textwidth}}{2048 channels, 2 layers per block, head dimension 64} \\
    Deep block & \multicolumn{2}{p{0.68\textwidth}}{Causal left-to-right LLM block} \\
    Soft-clip / dequantization & \multicolumn{2}{p{0.68\textwidth}}{Soft-clip $4.0$; dequantization noise $\sigma_{\text{dq}}=0.3$} \\
    Loss weights & \multicolumn{2}{p{0.68\textwidth}}{$\lambda_{\text{NF}}=1.0$, $\lambda_{\text{CE}}=1.0$} \\
    Prompt dropout & $p_{\text{drop}}=0.1$ at batch level & $p_{\text{drop}}=0.1$ per sample \\
    Latent-to-hidden projector & \texttt{flow\_projector} plus $W_{\text{dec}}$ & Shared \texttt{flow\_projector} \\
    Prefix marker(s) & \texttt{BOT}, \texttt{AE} & \texttt{BOT} \\

    Stage-1 frozen modules & Backbone and shallow blocks & Backbone and shallow blocks \\
    Stage-2 trainable modules & \multicolumn{2}{p{0.68\textwidth}}{All model parameters except the VAE encoder} \\
    Optimizer & \multicolumn{2}{p{0.68\textwidth}}{AdamW, $(\beta_1,\beta_2)=(0.9,0.95)$, weight decay $0.01$, gradient clip $1.0$} \\
    Learning rate / warmup & \multicolumn{2}{p{0.68\textwidth}}{Stage~1: $1\!\times\!10^{-4}$ with 100 warmup steps; Stage~2: $5\!\times\!10^{-5}$ with 200 warmup steps} \\
    Schedule & \multicolumn{2}{p{0.68\textwidth}}{Cosine decay} \\
    Batch configuration & \multicolumn{2}{p{0.68\textwidth}}{Per-device batch 2 / 1 and gradient accumulation 4 / 8 for Stage~1 / Stage~2} \\
    Hardware & \multicolumn{2}{p{0.68\textwidth}}{4 / 8 NVIDIA B200 GPUs for Stage~1 / Stage~2} \\
    Epochs / Stage-1 budget & \multicolumn{2}{p{0.68\textwidth}}{Stage~1: 1 epoch over 100,000 samples; Stage~2: 2 epochs} \\
    Precision / sharding & \multicolumn{2}{p{0.68\textwidth}}{bfloat16; FSDP \texttt{hybrid\_shard} with activation checkpointing} \\
    Random seed & \multicolumn{2}{p{0.68\textwidth}}{HuggingFace Trainer default when unset in the configuration} \\
    \bottomrule
  \end{tabular}
\end{table*}

\subsection{Inference Configuration}
\label{ssec:inference}

For the main benchmark evaluations, inference is split into two phases per
evaluation example. Phase~1
generates latent thoughts via the NF reverse pass; Phase~2 produces an
answer by feeding the latent prefix to the backbone as soft-token
\texttt{input\_embeds}. The speed ablation in
Table~\ref{tab:inference_efficiency_vllm} uses vLLM answer decoding as stated
there; this subsection describes the main benchmark decoding pipeline.

\paragraph{Phase~1: latent generation (PyTorch).}
We sample $z\sim\mathcal{N}(0, T_z^2 I)\in\mathbb{R}^{N\times D}$, prefix
the prompt embeddings with the model's boundary token (\texttt{BOT} for
both variants; the dual-path model adds \texttt{AE} only on the CE path),
and run an autoregressive NF reverse pass through the LLM with KV-cache
reuse: at each step $t$, the LLM hidden at the latest position is mapped
through \texttt{output\_head} to obtain $(\sigma_t, \mu_t)$, the next
latent is computed as $x_t = \sigma_t z_t + \mu_t$, the projected latent
$\mathrm{flow\_proj}(x_t)$ is appended, and the LLM is advanced by one
position over the cached keys/values. After $N=64$ steps, the deep
permutation is inverted; for the dual-path model we additionally invert
the five shallow MetaBlocks to recover raw latents in encoder space,
whereas the unified model keeps the latent in $\tilde x$ space (post-shallow,
post-permute) since the LLM was trained to read it there. Classifier-free
guidance is applied in $(\sigma,\mu)$ space following STARFlow~\cite{gu2025starflow}.

\paragraph{Phase~2: answer decoding (vLLM).}
The full backbone weights are extracted from the trained checkpoint and
loaded into an offline \texttt{vllm.LLM} engine with
\texttt{enable\_prompt\_embeds=True} (required for embedding inputs).
For each problem we feed the soft prefix
\(
[\,\mathrm{prompt\_embeds};\,\texttt{marker};\,\mathrm{proj}(\mathrm{latents})\,]
\)
through the engine's \texttt{prompt\_embeds} interface and sample answer
tokens. The marker and projector are $\texttt{AE}$/$W_{\text{dec}}$ for
the dual-path model and $\texttt{BOT}$/$\mathrm{flow\_projector}$ for the
unified model.

\paragraph{Decoding hyperparameters.}
For each benchmark problem, we draw 16 independent samples and report pass@1 as
the average single-sample correctness over the 16 candidates.
Each sample uses an independent NF noise draw
(no fixed seed at inference).
For Table~\ref{tab:main_results}, we use this standard $n=16$ sampling
configuration unless otherwise specified. For Figure~\ref{fig:passk}, we run a
separate pass@$k$ scaling evaluation with up to 128 sampled candidates per
problem. The pass@1 endpoints in that scaling run are reported from the same
candidate pool used for the pass@$k$ curves and are not used to compute
Table~\ref{tab:main_results}.
Decoding uses temperature $T_a = 0.6$, $\text{top-}p = 1.0$, and a
maximum of $1024$ new tokens; NF sampling temperature and CFG strength are
\emph{variant-specific} and were tuned on a held-out validation slice:
\begin{itemize}
  \item \textbf{Dual-path:} $T_z = 1.0$, $w_{\text{CFG}} = 3.5$.
  \item \textbf{Unified-path:} $T_z = 0.9$, $w_{\text{CFG}} = 1.0$ ; a
        moderately under-temperatured NF sampler is preferred for this
        model.
\end{itemize}

\subsection{Execution-Guided RL Details}
\label{app:rl_details}

\paragraph{Overview.}
We further apply execution-guided RL to the supervised \methodname{} (Unified)
checkpoint. During RL, the VAE encoder/decoder shell, shallow flow blocks,
flow projector, affine output head, and auxiliary latent-control embeddings are
kept frozen. We update only the shared backbone LLM, which is used both as the
autoregressive block of the normalizing flow and as the answer decoder. Thus,
the same trainable parameters receive reward-driven updates from both the
latent-trajectory policy and the answer-token policy.

\paragraph{Policy factorization.}
Each rollout consists of a latent trajectory
$\tilde{x}=(\tilde{x}_1,\ldots,\tilde{x}_N)$ with $N=64$, followed by an
answer token sequence $y=(y_1,\ldots,y_L)$. We view both the continuous latent
trajectory and the answer tokens as actions. 
The latent density is computed by inverting the affine transformation predicted
by the flow block at each latent position. For the realized latent
$\tilde{x}_t$, the model predicts affine parameters $(x_a^t,x_b^t)$ and
recovers the corresponding Gaussian noise as
\begin{equation}
z_t=\frac{\tilde{x}_t-x_b^t}{x_a^t}.
\end{equation}
The latent log-probability is then
\begin{equation}
\log p_\theta(\tilde{x}_t\mid p,\tilde{x}_{<t})
=
-\frac{1}{2}\|z_t\|_2^2
-
\sum_{d=1}^{D}\log |x_{a,d}^t|,
\end{equation}
up to the constant Gaussian normalization term, which cancels in the PPO
ratio. Since the shallow flow blocks are frozen and the action is defined in
post-shallow latent space, their log-determinants are constant with respect to
the trainable parameters and are omitted from the RL objective.

\paragraph{RL objective.}
For each prompt, we sample a group of $n=8$ rollouts and compute GRPO
advantages from their execution rewards:
\begin{equation}
\hat{A}_i
=
\frac{R_i-\mu}{\sigma+\varepsilon_\sigma},
\end{equation}
where $\mu$ and $\sigma$ are the group mean and standard deviation. The final
RL loss combines a standard token-level PPO objective with a sequence-level
latent PPO objective:
\begin{equation}
\mathcal{L}_\theta
=
\mathcal{L}^{\mathrm{tok}}_\theta
+
\mathcal{L}^{\mathrm{lat}}_\theta .
\end{equation}
The token term is computed on answer tokens only. The latent term applies the
same clipped PPO surrogate to the likelihood ratio of the complete latent
trajectory. Specifically, we define the per-step latent log-ratio as
\begin{align}
\delta_{i,t}^{\mathrm{lat}}
&=
\log p_\theta(\tilde{x}_{i,t}\mid p_i,\tilde{x}_{i,<t})
\nonumber \\
&\quad
-
\log p_{\theta_{\mathrm{old}}}(\tilde{x}_{i,t}\mid p_i,\tilde{x}_{i,<t}),
\\
\Delta_i^{\mathrm{lat}}
&=
\sum_{t=1}^{N}\delta_{i,t}^{\mathrm{lat}},
\\
r_i^{\mathrm{lat}}
&=
\exp\!\left(
\mathrm{clip}(\Delta_i^{\mathrm{lat}},-20,20)
\right).
\end{align}
Then the latent PPO term is
\begin{align}
\mathcal{L}^{\mathrm{lat}}_\theta
&=
-\mathbb{E}_i
\Big[
\min \Big(
r_i^{\mathrm{lat}}\hat{A}_i,\,
\bar{r}_i^{\mathrm{lat}}\hat{A}_i
\Big)
\Big]
\nonumber \\
&\quad
+
\beta_{\mathrm{KL}}\mathcal{K}^{\mathrm{lat}}_i ,
\label{eq:latent_ppo}
\end{align}
where
\begin{equation}
\bar{r}_i^{\mathrm{lat}}
=
\mathrm{clip}(r_i^{\mathrm{lat}},1-\varepsilon,1+\varepsilon).
\end{equation}
Here $\varepsilon=0.2$, and $\mathcal{K}^{\mathrm{lat}}_i$ is a sampled
log-ratio penalty between the actor and the frozen reference policy on the
same realized latent trajectory. We use the same KL coefficient for the token
and latent terms.

\paragraph{Optimization and evaluation.}
We train for $150$ RL steps with AdamW using learning rate $3\times 10^{-6}$,
batch size $64$ prompts, and $8$ rollouts per prompt. The KL coefficient is
$\beta_{\mathrm{KL}}=10^{-3}$, and the PPO clip range is $0.2$. During RL
rollout, we use answer temperature $0.9$ and latent temperature $1.0$.

After RL, we evaluate using the same decoding protocol as the supervised
\methodname{} models: answer temperature $0.6$, NF temperature $0.9$,
classifier-free guidance scale $1.0$, and $16$ samples per problem for
HumanEval, HumanEval+, MBPP, and MBPP+, LiveCodeBench v6.

\section{Additional Analysis}
\label{app:additional_analysis}

\subsection{Backbone Drift Diagnostic}
\label{app:backbone_drift}

To further diagnose the effect of the frozen-backbone warm-up stage, we
compare the default two-stage curriculum against a stage-2-only variant
that directly starts full-parameter joint training from randomly
initialized flow components. We analyze both early training dynamics and
the final backbone drift relative to the original Qwen3-8B-Base
checkpoint.

\paragraph{Early training dynamics.}
The two runs reach similar final optimization losses, but they differ
substantially at the beginning of joint training. In the default
curriculum, the flow-side components have already been trained while the
backbone is frozen. At the start of stage 2, the model has
$L_{\mathrm{NF}} \approx -0.42$ and log-determinant approximately
$-0.92$, indicating that the shallow flow blocks have moved away from an
identity map and already provide a nontrivial density model over the
latent targets.

By contrast, the stage-2-only run starts joint training with
$L_{\mathrm{NF}} \approx 0.47$ and log-determinant close to zero. This is
close to a random Gaussian-NLL baseline and suggests that the shallow
blocks initially behave almost like an identity map. Its initial
gradient norm is also substantially larger than that of the warm-started
run, $1.96$ versus $0.96$. These early dynamics suggest that, without
the warm-up stage, the pretrained backbone is immediately exposed to
large gradients from a poorly calibrated NF branch.

\begin{table}[t]
\centering
\caption{
Backbone drift relative to the original Qwen3-8B-Base checkpoint.
The overall drift difference is small, but the stage-2-only run shows
substantially larger drift in the final transformer layer.
}
\scriptsize
\setlength{\tabcolsep}{2.5pt}
\begin{tabularx}{\linewidth}{X c c c}
\toprule
Metric & Stage 1+2 & Stage-2-only & Diff. \\
\midrule
Overall rel-L2 & 0.1574 & 0.1599 & $1.016{\times}$ \\
Overall $1-\cos$ & 0.01237 & 0.01278 & $1.033{\times}$ \\
\midrule
Layer 0 rel-L2 & 0.142 & 0.144 & $+1\%$ \\
Layer 18 rel-L2 & 0.144 & 0.144 & $0\%$ \\
Layer 35 rel-L2 & 0.121 & 0.143 & $+18\%$ \\
Embed/head/norm rel-L2 & 0.261 & 0.261 & $0\%$ \\
\bottomrule
\end{tabularx}
\label{tab:backbone_drift}
\end{table}

\paragraph{Final backbone drift.}
Table~\ref{tab:backbone_drift} shows that the stage-2-only model does
not simply drift uniformly farther from the pretrained model. Its global
relative L2 distance is only $1.6\%$ larger than the default curriculum,
and the global cosine drift increases by only $3.3\%$. However, the
layerwise pattern is much more concentrated: the final transformer layer
has $18\%$ larger relative L2 drift, while the embedding layer, the
middle layer, and the embedding/head/normalization parameters are nearly
unchanged.

The final transformer layer is the shared
representation that directly feeds both the LM head and the NF output
head. In the stage-2-only run, this layer is exposed to early NF
gradients before the flow projector, shallow blocks, and output head have
learned a meaningful latent-density interface. As a result, the largest
additional movement appears exactly at the layer most directly coupled
to the randomly initialized NF branch. The warm-up stage mitigates this
effect by first stabilizing the flow-side components while the backbone
is frozen, so that the subsequent joint-finetuning stage sends a more
calibrated NF learning signal into the pretrained backbone.

\subsection{Pairwise Output Diversity}
\label{app:pairwise_diversity}

A key goal of latent-CoT sampling is not only to generate multiple
lexically different completions, but to explore genuinely different
algorithmic solutions. We therefore measure within-prompt output
diversity among passing programs under matched sampling budgets. For
four randomly selected HumanEval problems
(HumanEval/154, HumanEval/124, HumanEval/38, and HumanEval/123), we draw
64 samples per problem from \methodname{} and from the Qwen3-8B-Base
teacher. For \methodname{}, we use the same decoding configuration as
the main evaluation, with NF temperature $\tau_z=1.0$ and answer
temperature $T=0.6$; for the base model, we use the same answer
temperature $T=0.6$ and the same wrapped prompt. We keep only samples
that pass the HumanEval unit tests, and then uniformly sample $k=8$
passing programs per problem without replacement. To avoid reporting a
fortuitous subset, we repeat this sub-sampling over 64 random seeds and
visualize the seed whose mean intra-prompt similarity is closest to the
across-seed median.

\begin{figure*}[t]
\centering
\includegraphics[width=\textwidth]{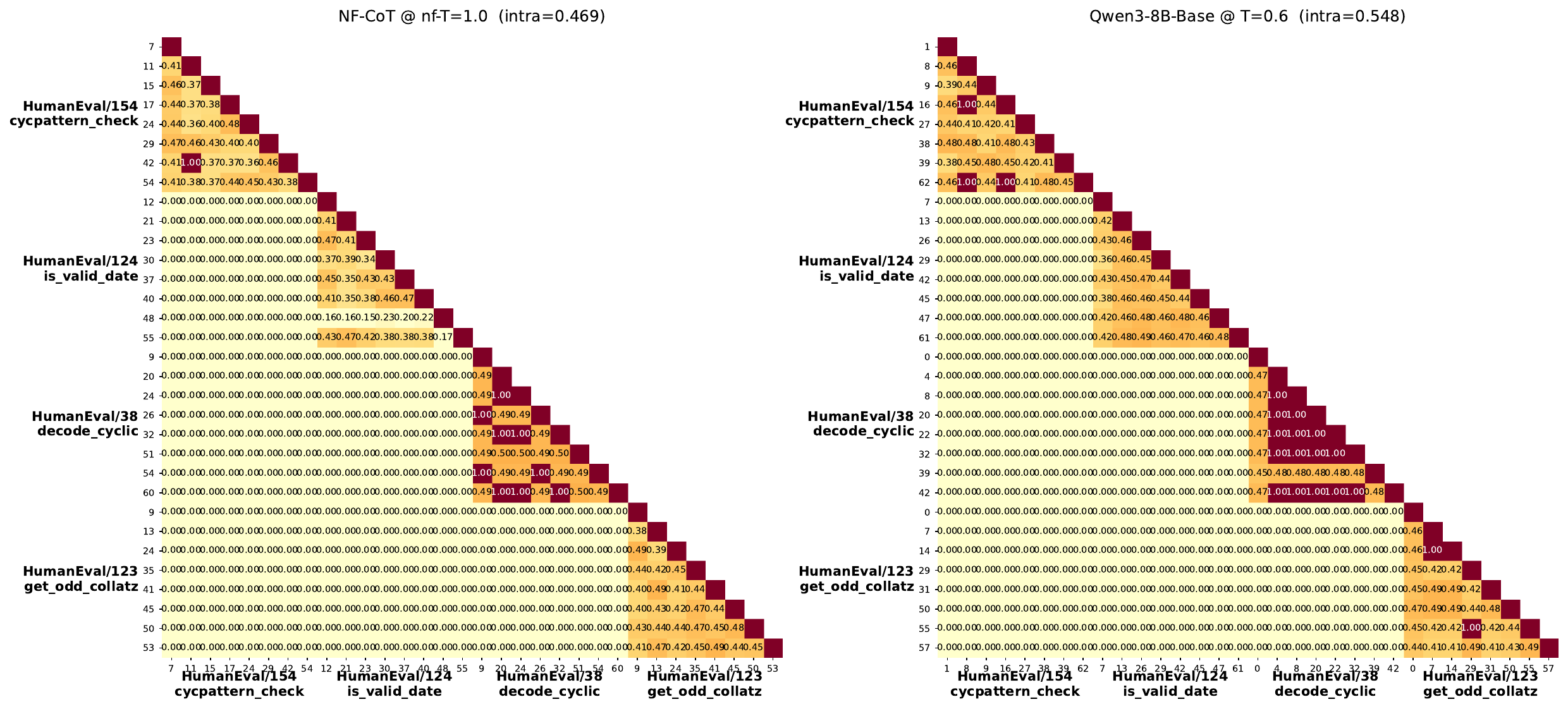}
\caption{
Pairwise output similarity among passing HumanEval programs. Each panel
contains $32=4\times 8$ passing samples from four HumanEval problems,
with intra-task blocks on the diagonal and cross-task entries set to
zero to isolate within-prompt structure. Similarity is computed using a
structure-aware AST metric; lower values indicate higher diversity.
\methodname{} achieves lower mean intra-prompt similarity than the
Qwen3-8B-Base teacher, indicating more structurally diverse passing
solutions under the same answer temperature.
}
\label{fig:pairwise_diversity}
\end{figure*}

Surface-level similarity metrics can be misleading in this setting.
Embedding cosine on raw generations, character $n$-gram similarity, and
edit distance are heavily affected by natural-language reasoning text,
variable names, formatting, and whitespace. At temperature $T=0.6$, the
base model often produces many lexically different paraphrases of the
same underlying algorithm, which can spuriously inflate apparent
diversity under token-level metrics. We therefore use a structure-aware
metric defined on the entry-point function body:
\[
s(a,b)
=
\frac{1}{2}
\cos\!\left(\phi_{\mathrm{2g}}(a),\phi_{\mathrm{2g}}(b)\right)
+
\frac{1}{2}\mathbb{I}\!\left[h(a)=h(b)\right].
\]
Here, $\phi_{\mathrm{2g}}(\cdot)$ is the count vector of parent-child
AST-node-type 2-grams, which captures control-flow and syntactic
patterns such as loop updates, list construction, calls, and conditional
branches. The function $h(\cdot)$ is the hash of a canonicalized AST
after local variables are $\alpha$-renamed to canonical identifiers and
docstrings/comments are stripped. The cosine term gives a smooth measure
of structural similarity between non-identical programs, while the hash
term pins truly identical canonical programs to similarity $1$. Both
terms are invariant to variable renaming, formatting, and comments.

Figure~\ref{fig:pairwise_diversity} shows that \methodname{} produces
more structurally diverse passing solutions than the base model. The
mean intra-prompt similarity drops from $0.548$ for Qwen3-8B-Base to
$0.469$ for \methodname{}, a relative reduction of approximately
$14\%$. The improvement is consistent across all four tasks: the
intra-prompt similarity decreases by $0.017$ on HumanEval/154, $0.156$
on HumanEval/124, $0.007$ on HumanEval/38, and $0.002$ on
HumanEval/123. The largest difference appears on HumanEval/124
(\texttt{is\_valid\_date}), where the base model largely collapses onto
a canonical split-and-branch implementation, whereas \methodname{}
samples a wider mixture of structurally different solutions, including
regular-expression-based, exception-handling-based, and
date-library-based implementations.

These results suggest that the latent flow does more than induce
surface-level variation. The base model is already lexically diverse
under temperature sampling, but its structural distribution remains
peaked around a small number of common implementation templates.
\methodname{} injects stochasticity upstream of token generation by
sampling from the latent prior $z\sim\mathcal{N}(0,\tau_z^2 I)$, so
different latent trajectories can steer decoding toward different
algorithmic regions before answer tokens are generated. As a result,
latent sampling changes which solution strategy the model follows,
rather than merely changing how the same strategy is verbalized or
formatted. We use the same decoding configuration as our main
evaluation, with $\tau_z=1.0$ and answer temperature $T=0.6$, so the
diversity comparison is aligned with the reported evaluation setting
rather than tuned specifically for this analysis.

\end{document}